\documentclass[10pt,twocolumn,letterpaper]{article}

\usepackage{iccv}
\usepackage[dvipsnames]{xcolor}

\definecolor{R1color}{rgb}{0,0.56,0}
\newcommand{\R}[1]{{%
    \textbf{%
        \ifstrequal{#1}{1}{\textcolor[rgb]{0,0.56,0}{R#1}}{%
        \ifstrequal{#1}{2}{\textcolor{blue}{R#1}}{%
        \ifstrequal{#1}{3}{\textcolor{magenta}{R#1}}{%
        \ifstrequal{#1}{4}{\textcolor{teal}{R#1}}{%
                           \textcolor{cyan}{R#1}%
        }}}}%
    }%
}}

\usepackage{algorithm}
\usepackage{algorithmic}

\usepackage{bm}
\renewcommand{\vec}[1]{\bm{#1}}

\def\paperID{6985}
\def\confName{ICCV}
\def\confYear{2025}
\def\paperTitle{DAViD: Modeling Dynamic Affordance of 3D Objects \\ Using Pre-trained Video Diffusion Models}

\def\authorBlock{
    Hyeonwoo Kim \qquad
    Sangwon Baik \qquad
    Hanbyul Joo \\
    [0.4em]
    Seoul National University \\
    [0.25em]
    {\tt\small \href{https://snuvclab.github.io/david/}{\color{magenta}{https://snuvclab.github.io/david/}}}
}

\usepackage{times}
\usepackage{microtype}
\usepackage{epsfig}
\usepackage[skip=5pt]{caption}
\usepackage{float}
\usepackage{placeins}
\usepackage{color, colortbl}
\usepackage{stfloats}
\usepackage{enumitem}
\usepackage{tabularx}
\usepackage{xstring}
\usepackage{cuted} %
\usepackage{multirow}
\usepackage{xspace}
\usepackage{url}
\usepackage{mathtools}
\usepackage[hang,flushmargin]{footmisc}
\usepackage[normalem]{ulem}  %
\usepackage{graphicx}
\usepackage{amsmath}
\usepackage{amssymb}
\usepackage{comment}
\usepackage[title]{appendix}
\usepackage{cancel}
\usepackage{bbm}
\usepackage{afterpage}
\usepackage{algorithm}
\usepackage{algorithmic}
\usepackage{bm}
\usepackage{textcomp}
\usepackage{gensymb}

\newfloat{sepfigure}{p}{lop}
\floatname{sepfigure}{Figure}

\usepackage{indentfirst}

\definecolor{iccvblue}{rgb}{0.21,0.49,0.74}
\usepackage[pagebackref,breaklinks,colorlinks,allcolors=iccvblue]{hyperref}
\title{\paperTitle}
\author{\authorBlock}
\begin{document}
\maketitle
\begin{strip}
    \centering
    \vspace{-46pt}
    \captionsetup{type=figure}
    \includegraphics[width=\textwidth]{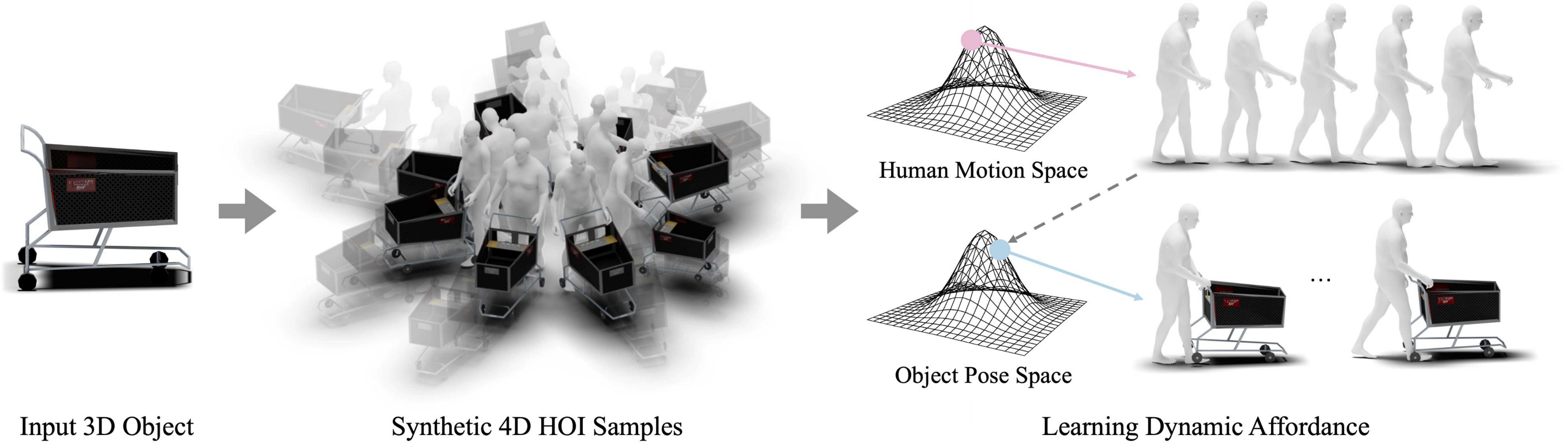}
    \captionof{figure}{\textbf{DAViD for Learning Dynamic Affordance.} Given an input 3D object (left), our method, DAViD, learns Dynamic Affordance to model the dynamic patterns that occur during HOI (right). To build it, we present a framework to synthesize diverse 4D HOI samples (middle) by leveraging a pre-trained video diffusion model.}
    \label{fig:teaser}
    \vspace{-5pt}
\end{strip}

\begin{abstract}
Modeling how humans interact with objects is crucial for AI to effectively assist or mimic human behaviors.
Existing studies for learning such ability primarily focus on static human-object interaction (HOI) patterns, such as contact and spatial relationships, while dynamic HOI patterns, capturing the movement of humans and objects over time, remain relatively underexplored.
In this paper, we present a novel framework for learning Dynamic Affordance across various target object categories.  
To address the scarcity of 4D HOI datasets, our method learns the 3D dynamic affordance from synthetically generated 4D HOI samples. 
Specifically, we propose a pipeline that first generates 2D HOI videos from a given 3D target object using a pre-trained video diffusion model, then lifts them into 3D to generate 4D HOI samples.
Leveraging these synthesized 4D HOI samples, we train DAViD, our generative 4D human-object interaction model, which is composed of two key components: (1) a human motion diffusion model (MDM) with Low-Rank Adaptation (LoRA) module to fine-tune a pre-trained MDM to learn the HOI motion concepts from limited HOI motion samples, (2) a motion diffusion model for 4D object poses conditioned by produced human interaction motions.
Interestingly, DAViD can integrate newly learned HOI motion concepts with pre-trained human motions to create novel HOI motions, even for multiple HOI motion concepts, demonstrating the advantage of our pipeline with LoRA in integrating dynamic HOI concepts.
Through extensive experiments, we demonstrate that DAViD outperforms baselines in synthesizing HOI motion.
\end{abstract}
    
\section{Introduction}
\label{sec:intro}

Humans effortlessly understand how to effectively use objects to achieve their goals within a given environment.
This knowledge, known as affordance originally introduced by Gibson~\cite{affordance}, encapsulates a comprehensive understanding of the inherent ways humans interact with objects.
Modeling these patterns is crucial for AI systems to assist or mimic human behavior, leading to extensive research in AI fields~\cite{NCHO, affordance_classification_02, affordance_classification_03, affordance_classification_04, affordance_classification_05, affordance_classification_06, affordance_classification_07, affordance_detection_01, affordance_detection_02, affordance_detection_03, affordance_detection_04, affordance_detection_05, affordance_detection_06, affordance_detection_07, affordance_detection_08, affordance_reasoning_01, affordance_reasoning_02, affordance_reasoning_03, affordance_reasoning_04, affordance_reasoning_05, affordance_reasoning_06, affordance_reasoning_07, affordance_reasoning_08, affordance_reasoning_09, affordance_reasoning_10, affordance_segmentation_1, affordance_segmentation_2, affordance_segmentation_3, affordance_segmentation_4}.
However, existing studies on affordance learning primarily focus on static human-object interaction (HOI) patterns. For example, 
some approaches address affordances by inferring contact (or interaction) region between humans and objects in 2D images~\cite{LOCATE, Robo-Affordance}, 3D humans~\cite{DECO, LEMON}, and 3D objects~\cite{IAGNet, LEMON}.
Others~\cite{CHORUS, ComA} represent affordances as spatial distributions aggregated from static 3D HOI samples.
While recent methods generate human motions including HOI~\cite{OMOMO, InterDiff, HOIDiff, CHOIS, InterDreamer, HOIfHLI, lama} and Human-Scene Interaction (HSI)~\cite{AffordMotion, TesMo, HOIfHLI}, they often focus on limited actions such as standing, sitting, or picking, without focusing on rich, dynamic nature of object-specific affordances in human motion.

\begin{figure}[t]
\centering
\includegraphics[width=\columnwidth]{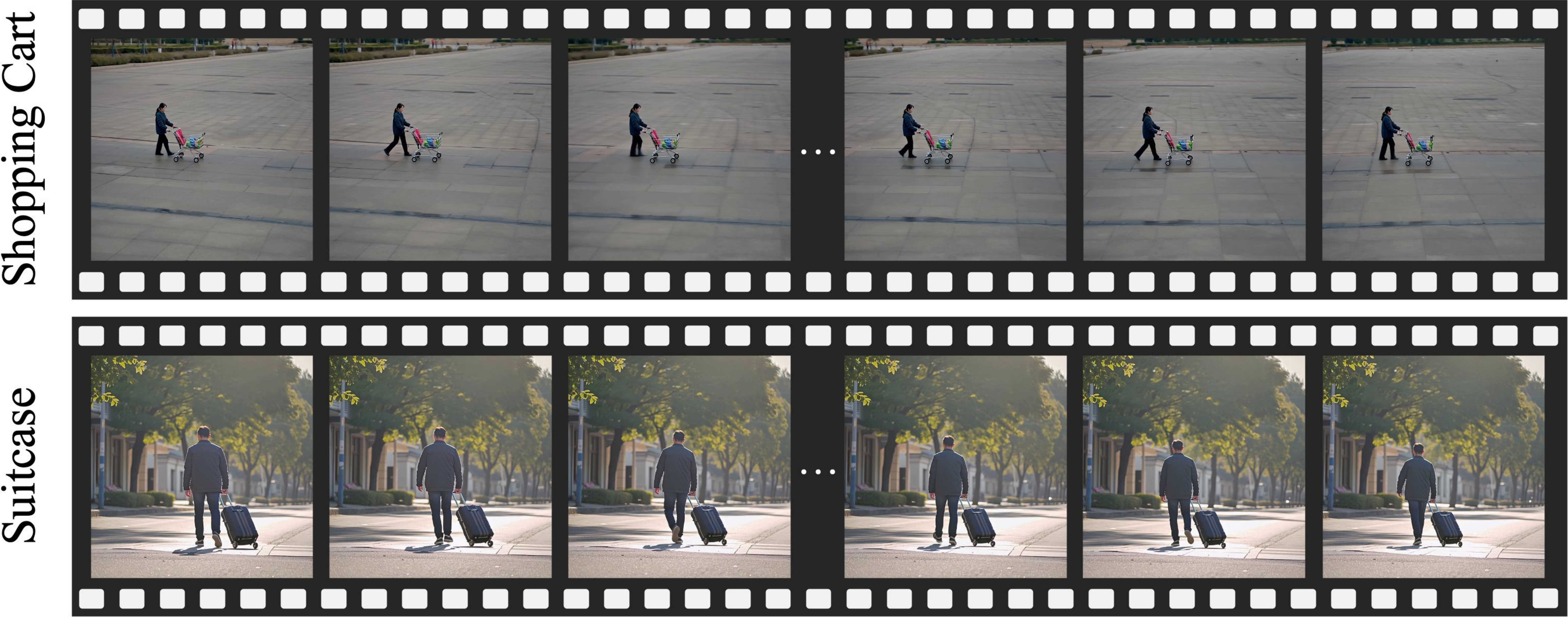}
\caption{\textbf{Dynamic Patterns in HOI.} People show dynamic patterns during HOI. Pre-trained video diffusion models have the knowledge of the dynamic patterns.} 
\label{fig:motivation}
\vspace{-15pt}
\end{figure}

In this paper, we present a novel framework for learning Dynamic Affordance across diverse and unbounded target object categories, without the need for laborious 3D HOI data capture.
Our approach is motivated by the observation that human-object interactions involve not only static properties (e.g., contact regions or grasping patterns) but also object-specific characteristic movements over time. 
For instance, as shown in Fig.~\ref{fig:motivation}, humans naturally push a shopping cart and pull a suitcase while walking forward, yet rarely lift their handles vertically.
Existing studies, which mainly focus on static affordance features, fail to capture these dynamic interaction patterns.
To address this, we present DAViD, a generative motion model for modeling Dynamic Affordance that learns the dynamic and object-specific 4D HOI patterns. 
To handle the vast diversity of dynamic HOI patterns, our method considers each object-specific 4D HOI pattern as a distinct motion concept, adapting a pre-trained motion diffusion model (MDM)~\cite{MDM} via Low-Rank Adaptation (LoRA)~\cite{LoRA}. Once human motions are generated by our MDM with LoRA, we further synthesize object motions conditioned by human motion using a score-based diffusion model, where we introduce a novel sampling technique, named Temporal Guidance Sampling (TGS), for enhancing motion consistency and fine details.
Notably, our approach not only models dynamic affordances from limited data but also seamlessly blends newly learned HOI motion concepts with those already learnt in the pre-trained model, enabling zero-shot generalization, even across multiple HOI concepts.

To train DAViD, we also present a pipeline to synthetically produce realistic 4D HOI samples for a target object category, enabling effective model training without requiring expensive real-world 3D data collection.
Most existing dynamic HOI learning~\cite{PhysHOI, CHAIRS, COUCH} or mimicking approaches~\cite{InterMimic} rely on the 3D datasets captured in lab-controlled environments~\cite{BEHAVE, InterCap, PROX, COINS, GRAB, ARCTIC, ParaHome}. This significantly limits the opportunity to learn diverse and distinctive dynamic affordance patterns across a broader range of object categories. 
Our key insight is that pre-trained video generation models inherently capture dynamic patterns of object usage.
As shown in Fig.~\ref{fig:motivation}, these models already know plausible dynamic interactions (\eg, pushing a shopping cart or pulling a suitcase).
Since it presents non-trivial challenges to leverage the synthetic 2D videos for learning 4D dynamic affordance, we propose a novel pipeline that synthesizes plausible 4D HOI samples for a given 3D object mesh. We achieve this by generating 2D videos aligned to the 3D object, and aligning the uplifted 4D human motion sequences with the object model, guided by synthetic video cues.

We validate the efficacy of our approach by learning Dynamic Affordance on 30 3D object categories, gathered from multiple sources~\cite{BEHAVE, InterCap, ShapeNet, SAPIEN, PartNet, ComA, SketchFab}.
We compare our DAViD with other HOI motion synthesis approaches~\cite{OMOMO, CHOIS, MDM} against public dataset~\cite{OMOMO}, demonstrating that our model outperforms competitors in generating HOI motion with plausible contact.
We further demonstrate the advantage of our pipeline, by generating results for a various previously unexplored target objects, and also validating its advantage in merging multiple HOI motion concepts with known human motion from pre-trained MDM~\cite{MDM}.

In summary, our main contributions are as follows: (1) We present a generative motion model, DAViD, to learn Dynamic Affordance; (2) We present a pipeline to synthetically generate diverse 4D HOI samples leveraging video diffusion models; (3) We address generating various novel HOI motions by fusing multiple, separately trained LoRAs with known human motions from a pre-trained model.
The results and our source code are publicly available.

\section{Related Work}
\label{sec:related}

\noindent\textbf{Reasoning Visual Affordance.}
Affordance is a concept defined by Gibson~\cite{affordance} as the set of possible interactions that an agent can perform within an environment.
As the concept narrows to the primary agent (or humans) and the specific interaction target (or objects), the area of learning affordance mainly focuses on HOI.
Specifically, followed by the intuition that the visual geometry of an object is closely related to its functionality, many studies leverage visual cues to reason about object's affordance.
The studies are subdivided into various areas, including affordance classification~\cite{affordance_classification_01, affordance_classification_02, affordance_classification_03, affordance_classification_04, affordance_classification_05, affordance_classification_06, affordance_classification_07}, affordance detection~\cite{affordance_detection_01, affordance_detection_02, affordance_detection_03, affordance_detection_04, affordance_detection_05, affordance_detection_06, affordance_detection_07, affordance_detection_08}, and affordance segmentation~\cite{affordance_segmentation_1, affordance_segmentation_2, affordance_segmentation_3, affordance_segmentation_4}. 
Such traditional studies infer affordance for the given context (mostly provided as images), as a form of action labels~\cite{affordance_action_label_01, affordance_action_label_02, affordance_action_label_03}, and contact (or interaction) regions in 2D~\cite{LOCATE, Robo-Affordance} or 3D~\cite{DECO, IAGNet}.
Later studies model static HOI patterns (\eg, contact) during interactions~\cite{3DAffordanceNet, CHORUS, ComA}, going beyond the understanding of individual HOI situations.
Recent studies synthesize human motion including HOI~\cite{InterDiff, HOIDiff, AffordMotion, TesMo, CHOIS, InterDreamer, HOIfHLI}, showing potential to learn dynamic HOI patterns, but they focus on a text-driven approach, generating consistent human motions regardless of object categories (\eg, dragging a chair, holding a backpack).
In contrast, our method focuses on learning rich, dynamic object-specific HOI patterns that are observable in real-world HOIs.

\noindent\textbf{Learning from Synthesized Data.} 
Generative models such as GANs~\cite{gan} and diffusion models are used as a strategy to address the problem of insufficient real-world data in many fields.
Starting from augmentation~\cite{perse, OOR, data_generate_diffusion_01, data_generate_diffusion_02, data_generate_diffusion_03, data_generate_diffusion_04}, synthetic datasets are widely used in vision area including segmentation~\cite{data_generate_segmentation_01, data_generate_segmentation_02, data_generate_segmentation_03}, classification~\cite{data_generate_classification_01, data_generate_classification_02, data_generate_classification_03}, shape reconstruction~\cite{data_generate_shape_reconstruction_01}, and neural rendering~\cite{data_generate_neural_rendering_01, data_generate_neural_rendering_02}.
While data synthesis is a powerful technique with broad applications, a key consideration when learning from generated data is its reliability compared to real-world data.
Recently, a few studies in the HOI field~\cite{CHORUS, ComA} leverage pre-trained 2D diffusion models to generate 2D HOI images, demonstrating the reliability of synthetic data to learn affordance knowledge. While following the approaches, we focus on learning dynamic patterns during HOI rather than learning static patterns.

\noindent\textbf{Learning Concepts in Diffusion Models.} 
Concept learning, also known as concept customization, is a strategy for extending the space of the pre-trained model to understand specific concepts (\eg, subjects, styles).
Some studies tune the embeddings~\cite{textual_inversion, P+} corresponding to a given concept in a pre-trained diffusion model, while others~\cite{DreamBooth, custom_diffusion} tune both the model weights and embeddings together.
Recently, LoRA~\cite{LoRA}, originally used in large language models (LLMs), shows impressive results in producing images well aligned to the given concept~\cite{LoRA_image}, leading to an increase of attempts~\cite{lora_example_01, lora_example_02, lora_example_03} to use LoRA for concept learning.
Such concepts can be provided in various forms, including images~\cite{concept_image_01, concept_image_02, concept_image_03, concept_image_04}, videos~\cite{concept_video_01, lora_example_02}, and audios~\cite{concept_audio_01}.
However, concept learning in fields beyond images, video, and audio is relatively less explored.
We apply LoRA to the human motion space of pre-trained MDM~\cite{MDM} to learn the dynamic patterns displayed in HOI as a single concept, extending the original human motion space in terms of HOI.

\section{Method}
\label{sec:method}

Our approach aims to learn the Dynamic Affordance of a target object from synthetically generated 2D videos, which are produced by a video diffusion model~\cite{kling}. In the first stage, we generate diverse 4D HOI samples interacting with a given 3D object mesh by leveraging pre-trained video models (Sec.~\ref{subsec:dataset_generation}).
Then, our method models the collected 4D HOI samples into our DAViD, which consists of (1) a motion diffusion model along with LoRA, and (2) an object pose diffusion model.
The first module synthesizes feasible human motions for the target object, and the second module synthesizes object motions paired with previously generated human motion, using our novel sampling technique, Temporal Guidance Sampling (TGS) (Sec~\ref{subsec:dyna}).  
An overview of our method is shown in Fig.~\ref{fig:overview}.

\subsection{4D HOI Sample Generation}
\label{subsec:dataset_generation}
Our pipeline begins with a 3D object mesh input, and produces diverse 2D HOI videos showing how a person interacts with (or uses) the target object.

\noindent\textbf{2D HOI Video Synthesis.}
Given an input 3D object mesh, we produce multiple 2D HOI videos capturing human-object interactions with the object.
We first render the object from cameras with multiple viewpoints.
The intrinsic parameter is fixed to match the same perspective camera model and parameters of the world-grounded human mesh recovery method (GVHMR~\cite{gvhmr}) used in our later 3D lifting process, ensuring a consistent camera setup. 
For the extrinsic parameters, we consider diverse viewpoints based on the object's type, either ground-placed or portable objects.
For relatively stationary objects (\eg, cart), we consider horizontal viewpoint variations centered on the object, while for the portable objects (\eg, trumpet) we sample viewpoints within a specified range from all view angles. The object type is automatically inferred via a vision-language model~\cite{gpt4o}, by querying with the rendered image with a prompt: \textit{``Is the object plausibly on the ground while being used?''}. 

Once the object is rendered from each view, we synthesize a realistic human naturally interacting with it using an off-the-shelf pre-trained 2D image diffusion model~\cite{flux}. 
Unlike the previous similar approaches~\cite{Affordance-Insertion, Affordance-Diffusion, ComA} that use inpainting by providing a mask region to a model, we use ControlNet~\cite{ControlNet} to synthesize the entire image, including the human and the plausible background, by using edge cues of rendered object~\cite{canny} as the condition.
Our primary objective is to synthesize both plausible human-object interaction for the given object viewpoint and a realistic background, which is important for estimating dynamic cues and camera poses in a global coordinate.
After synthesizing HOI images from several views, we refine the results by filtering out low-quality or undesirable images. 
The filtering process can be performed efficiently with minimal human effort, requiring only about three minutes per category.

Finally, we synthesize corresponding 2D HOI videos ${\{\mathcal{V}_d\}_{d=1}^N}$ from each 2D HOI images ${\{\mathcal{I}_d\}_{d=1}^N}$ by using a pre-trained image-to-video generation model~\cite{kling} with the same text prompt used for generating images\footnote{As we synthesize several images at each camera view, the number of videos $N$ is not necessary the same as camera view number, $C$.}. 
Notably, our synthesized 2D HOI videos have key advantages over in-the-wild 2D real videos in learning dynamic affordance: (1) our HOI videos contain diverse views and interaction scenarios for the input 3D object, (2) we have precise camera parameters and 3D object pose information for the initial frame of each video. These are important factors for reliable 4D HOI sample generations.

\begin{figure*}[t]\centering
\includegraphics[width=\linewidth, trim={0 0 0 0},clip]{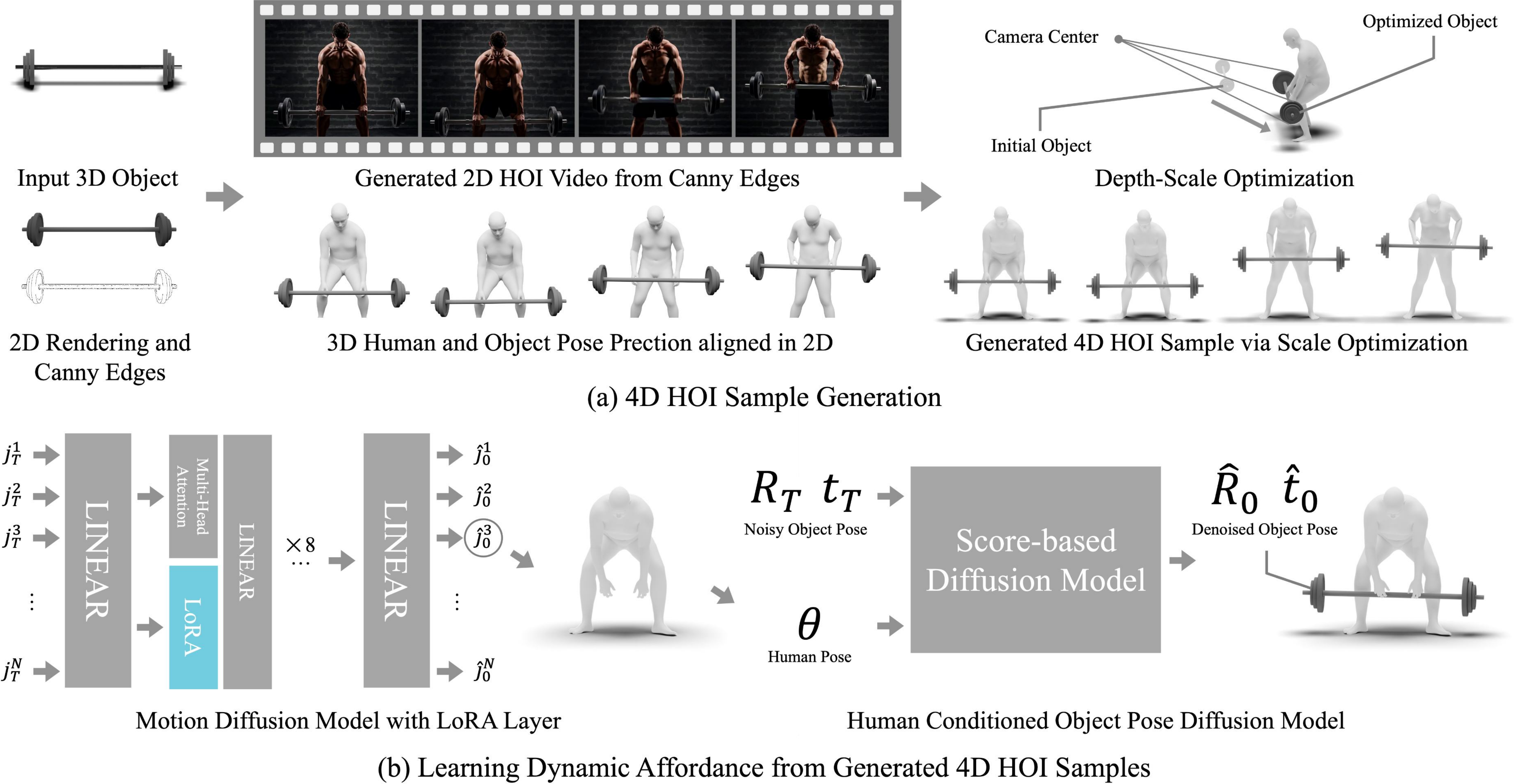}
\captionof{figure}{\textbf{Overview.} Our method consists of two parts: (1) Generating 4D HOI Samples and (2) Learning Dynamic Affordance from Generated 4D HOI Samples. First, we create 2D HOI videos, and generate 4D HOI samples with our uplifting pipeline. Then, the generated 4D HOI samples are used to train DAViD, learning the patterns of human motion and object pose.
}
\label{fig:overview}
\vspace{-10pt}
\end{figure*}

\noindent\textbf{Uplifting for 4D HOI Sample Generations. }
For each 2D HOI video $\mathcal{V}$~\footnote{we drop the video index $d$ for simplicity.}, we uplift 3D human motions and 3D object movements to generate 4D HOI samples.
Our key strategy is to uplift each component (human, object) by finding each motion relative to each camera, and co-locate them in the common camera coordinate shared by both components.

For the object side, we use the Perspective-n-Points (PnP)~\cite{PnP, EPnP} algorithm to find the relative camera pose with respect to the given 3D object mesh. 
Specifically, we aim to compute the extrinsic camera poses 
$\{ \vec{R}_{o}^{f}, \vec{t}_{o}^{f} \}_{f=1}^F$, where the one for the first frame $f=1$ is already known.
We first identify visible object mesh vertices $ \{ \mathbf{V}_{o} \}_{\text{vis}}$ via raycasting~\cite{raycast} for the first frame $f = 1$, where $ \mathbf{V}_{o} \in \mathbb{R}^3 $ is an object vertex. The 2D projections of the visible vertices at the first frame is denoted as $\{ \vec{v}_{o}^{f=1} \}_{\text{vis}}$, where $ \vec{v}_{o}^{f=1} \in \mathbb{R}^2$.
By applying an off-the-shelf 2D video tracker~\cite{cotracker, cotracker3}, we obtain the corresponding 2D points $\vec{v}_{o}^f$ in the remaining frames $f>1$.
As we have the correspondences between 3D object vertices and matched 2D points for every frame, we apply PnP~\cite{PnP, EPnP} to compute the camera poses at frame $f$ with respect to the object mesh in its canonical space:
\begin{equation}
     (\vec{R}_{o}^{f}, \vec{t}_{o}^{f}) = \text{PnP}\left( \{ \vec{v}_{o}^{f} \}_{\text{vis}}, \{ \vec{V}_{o} \}_{\text{vis}} \right).
\end{equation}

For the human side, we leverage world-grounded human motion recovery models, denoted as $\mathbf{F}_{\text{human}}$ to uplift human motions and corresponding camera motions into 3D from the synthesized 2D HOI videos. We use GVHMR~\cite{gvhmr}, except in the cases where human translation occurs without walking motions (\eg, riding a motorcycle). In such cases, we use TRAM~\cite{tram}, as GVHMR~\cite{gvhmr} typically fails to predict plausible camera motion under these conditions. Therefore, 
\begin{equation}
    \mathbf{F}_{\text{human}}(\mathcal{V}) = \{   \vec{\theta}^{f} ,  \vec{\beta}^{f} , \vec{\phi}^{f}, \vec{\tau}^{f},  \vec{R}_{h}^{f}, \vec{t}_{h}^{f}\}_{f=1}^F,
\end{equation}
where $\vec{\theta}^{f} \in \mathbb{R}^{21\times3}$, $\vec{\beta}^{f} \in \mathbb{R}^{10}$, $\vec{\phi}^{f} \in \text{SO(3)}$, $\vec{\tau}^{f} \in \mathbb{R}^{3}$ are the predicted SMPL-X~\cite{SMPLX} parameters of human pose, shape, root joint's rotation, and translation for $f$-th frame, respectively. 
The $\vec{R}_{h}^{f} \in \text{SO(3)}$, $\vec{t}_{h}^{f} \in \mathbb{R}^{3}$ are the extrinsic parameters of estimated camera poses. Note that all of the parameters are defined in a world coordinate for the human mesh space, which is different from object mesh space. 

Intuitively, the human and object at time $f$ are observed from the same camera, and thus we can simply co-locate them in the same space, by transferring each component to its camera coordinate, where the camera becomes the origin:
\begin{gather}
    \vec{\hat{V}}_{o}^{f} = \vec{R}_o^f \vec{V}_{o}^{f}  + \vec{t}_o^f, \\
    \vec{\hat{V}}_{h}^{f} = \vec{R}_h^f \vec{V}_{h}^{f}  + \vec{t}_h^f,
\end{gather}
where $\vec{V}_{h}^{f} \in \mathbb{R}^3$ is a human 3D vertex in its global coordinate after applying all SMPL parameters, and $\vec{\hat{V}}_{o}^{f}$ and $\vec{\hat{V}}_{h}^{f}$ are the human and object 3D vertices in the shared camera coordinate.
To this end, we align the object-side cameras to the human-side cameras, bringing cameras, humans, and objects into a unified coordinate system of human side.

\noindent \textbf{Resolving Depth-Scale Ambiguity.}
Although we uplift and place human and object movements in a common 3D space, there still exists the depth-scale ambiguity, since each components are defined in arbitrary object scales, as shown in depth-scale optimization part in Fig.~\ref{fig:overview}.
To resolve the ambiguity, we optimize the scale of the human ($s_h \in \mathbb{R}$) and object ($s_o \in \mathbb{R}$) in the camera coordinate at the first frame, using (1) a weak depth cue from the depth map, and (2) a contact cue from the object movement assumption. The objective function is defined as follows:
\begin{gather}
    \mathcal{L}_{\text{total}} = \lambda_h\mathcal{L}_h + \lambda_o\mathcal{L}_o + \lambda_{\text{HOI}}\mathcal{L}_{\text{HOI}} + \lambda_{\text{col}}\mathcal{L}_{\text{collision}},
\end{gather}
where $\lambda$s denote the weighting factors. The $\mathcal{L}_h$ and $\mathcal{L}_o$ are the depth consistency loss for human and object, defined as:
\begin{align}
    \mathcal{L}_h &= \sum_{i \in \text{vis}_{\text{h}}}  \left\lVert s_h \vec{\hat{V}}_{h,i}  - \vec{P}_{h,i} \right\rVert^2 \\
    \mathcal{L}_o &= \sum_{j \in \text{vis}_{\text{o}}} \left\lVert  s_o \vec{\hat{V}}_{o,j}  - \vec{P}_{o,j} \right\rVert^2,
\end{align}
where $\vec{\hat{V}}_{h,i}$ and $\vec{\hat{V}}_{o,i}$ are the $i$-th human and object vertices in the camera coordinate at the first frame respectively, and  $\vec{P}_{h, \text{3D}}$, $\vec{P}_{o, \text{3D}}$ are the corresponding 3D points by back-projecting 2D metric depths (computed by metric depth model~\cite{depth_pro}) into the camera coordinate. 
Note that the indices $i, j$ are from the set of visible vertex indices of the human and object at the first frame.

Additionally, the loss term of the weak HOI contact cue is defined as follows: 
\begin{gather}
    \mathcal{L}_{\text{HOI}} = \mathcal{D}_n\left( \{s_h \vec{\hat{V}}_{h,i} \},  \{ s_o \vec{\hat{V}}_{o,j} \} \right),
\end{gather}
where $\mathcal{D}_n$ is the function to compute the average distance of the $n$ closest points between two points sets. Intuitively, the cost term encourages contact between the human and object, assuming they are in contact at the first frame. Note that $n$ is a hyperparameter that varies across categories.
We apply the collision term to avoid undesired penetration using COAP~\cite{coap}. The object vertices are used as query points, reducing the collision between human and object in $\mathcal{L}_{\text{collision}}$.

\subsection{Learning Dynamic Affordance}
\label{subsec:dyna}
Based on the diverse dynamic 4D HOI samples we synthesized for a target 3D object, we build a generative HOI motion model, DAViD, to model dynamic affordance. Our DAViD consists of two key components: (1) a generative human motion model by fine-tuning a pre-trained MDM~\cite{MDM} with object-specific LoRA~\cite{LoRA} module, and (2) a score-based diffusion model for estimating object poses conditioned by generated human poses. For the score-based diffusion model, we introduce a novel inference-time sampling technique, TGS to enhance overall quality of motion.

\begin{figure*}[t]\centering
\vspace{-5pt}
\includegraphics[width=1.0\linewidth, trim={0 0 0 0},clip]{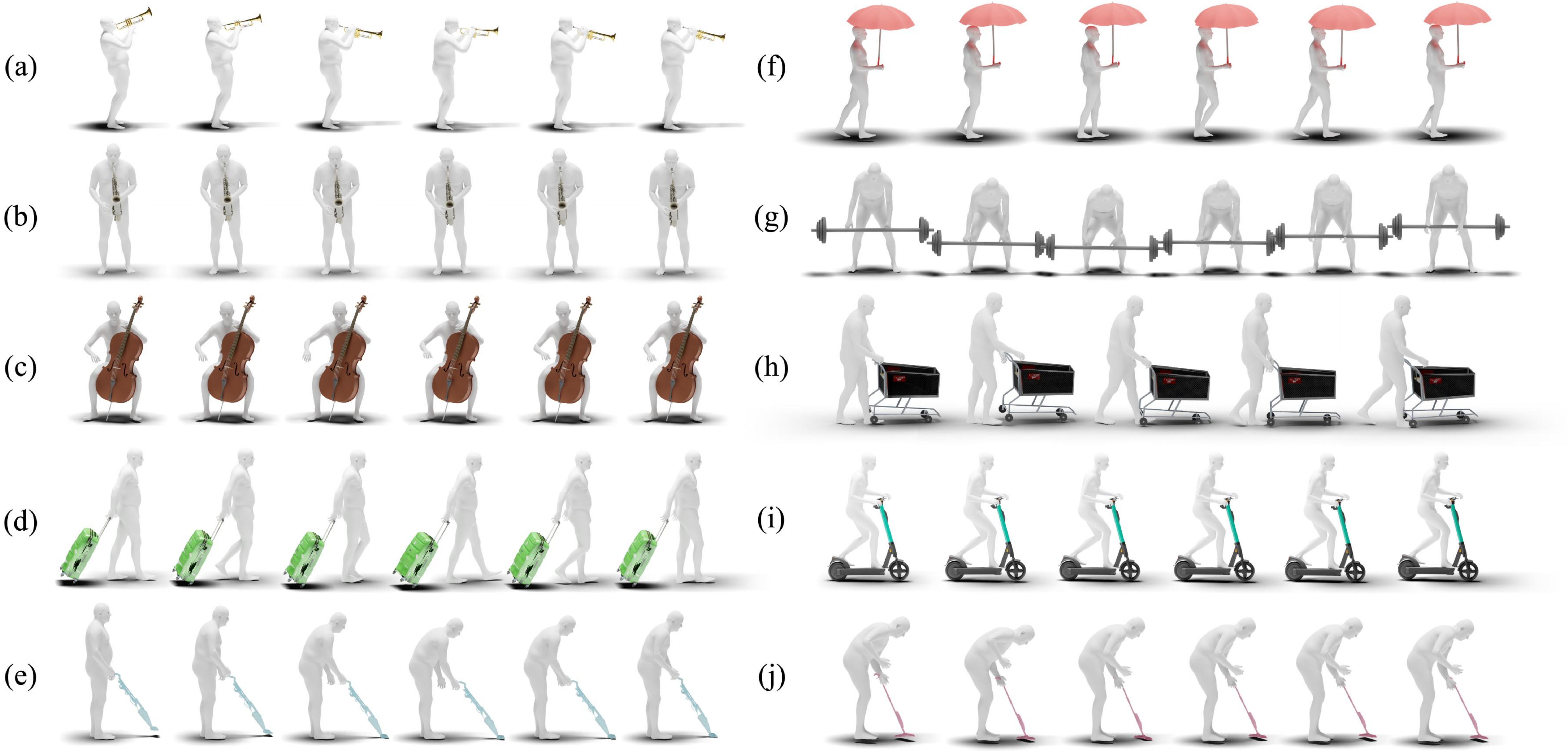}
\captionof{figure}{\textbf{
Qualitative Results. } DAViD models the dynamic patterns of both a human and an object for various object categories. 
We visualize HOI motions for various objects generated by our DAViD, where frames are visualized in temporal order. 
}
\label{fig:qual}
\vspace{-10pt}
\end{figure*}

\noindent\textbf{Human Motion Synthesis with LoRA for MDM. }
To capture human motion patterns interacting with a specific object, we insert a LoRA~\cite{LoRA} layer into a pre-trained MDM~\cite{MDM} model and train it for each object category.
By leveraging LoRA~\cite{LoRA}, our model can learn previously unseen human motion patterns with a limited amount of training samples, keeping the original motion synthesis quality of the pre-trained model.
Interestingly, we find that merging LoRA~\cite{LoRA} layers trained across different object categories enables our model to generate human motion interacting with multiple objects in a zero-shot manner (see Fig.~\ref{fig:concepts}~(c)).

For training, we extract only the human motion from our 4D HOI samples, excluding object motions. We preprocess the motion data following HumanML3D~\cite{HumanML3D}, and obtain the training data of human joints denoted as $\vec{j} \in \mathbb{R}^{F \times 22 \times 3}$.
The LoRA~\cite{LoRA} layer is then added into the multi-head attention layer of the transformer encoder block in text-conditioned MDM~\cite{MDM} as shown in Fig.~\ref{fig:overview}. Our model is trained with the following joint recovery loss function.
\begin{equation}
    \mathcal{L}(\Delta \phi) = \mathbb{E}_{\vec{j}_0, t} \| \vec{j}_0 - \mathcal{M}_{\phi + \Delta\phi}(\vec{j}_t, t, c) \|_2^2,
\end{equation}
where $\phi$ is the original weight of pre-trained MDM~\cite{MDM} denoted as $\mathcal{M}$, $c$ is the encoded text prompt using CLIP encoder~\cite{CLIP}, and $\Delta \phi$ is the weight of our LoRA~\cite{LoRA} module we want to obtain. For the training text prompt, we use the category name and annotate the additional hand index (left or right) for categories with hand-object interactions to improve controllability. See Supp. Mat. for more details.

\noindent\textbf{Object Motion Synthesis from Human Motion. }
After generating plausible human motion for a given 3D object, we synthesize a corresponding 3D object motion conditioned on the human motion.
To achieve it, we design our object motion module with two key parts: (1) a score-based diffusion model for modeling the distribution of object pose conditioned on the corresponding 3D human pose in HOI, (2) an inference method to estimate the final motion trajectory across a sequence of frames, with a novel sampling technique, named Temporal Guidance Sampling (TGS), for enhancing motion consistency and fine contact details.

To train the score-based diffusion model, we first normalize each frame of the generated 4D HOI sample by adjusting the translation so that the pelvis joint aligns with the origin, and rotate the root orientation to align with z-axis of the world coordinate.
The transformations are applied to both the human and the object.
A pair of the SMPL-X~\cite{SMPLX} body pose parameters $\vec{\theta} \in \mathbb{R}^{21\times 3}$ and the normalized object pose $\vec{p} \in \mathbb{R}^{9}$ (consists of 6D rotation and 3D translation) are used for training our diffusion model.
Using the pre-processed data samples, we train our score-based diffusion model $\Psi$ to output the score of the conditional object pose ($\vec{p}_t$) distribution for all timestep $t$:
\begin{gather}
    \Psi(\vec{p}, t | \vec{\theta}) = \nabla_{\vec{p}} \log p (\vec{p}_t | \vec{\theta}).
\end{gather}
As demonstrated in Pascal et al.~\cite{score_matching}, the objective can be achieved by minimizing the following loss function.
\begin{gather}
    \mathcal{L} = \mathbb{E}_{t \sim \mathcal{U}(\epsilon, 1)} \lambda(t) \mathbb{E}_{\vec{p}, t} \left\lVert \Psi(\vec{p}_t, t | \vec{\theta}) - \frac{\vec{p}_0 - \vec{p}_t}{\sigma(t)^2} \right\rVert_2^2 \\
    \sigma(t) = \sigma_{\text{min}} \left( \frac{\sigma_{\text{max}}}{\sigma_{\text{min}}} \right)^t,
\end{gather}
where $\sigma_{\text{min}} = 0.1$, $\sigma_{\text{max}} = 50$ are hyperparameters about noise perturbation.
For the detailed structure of our score model, we follow the architecture of GenPose~\cite{GenPose}.

Based on our conditional object pose diffusion model, we present an inference method to estimate the entire object motion sequence, rather than sampling object poses independently per frame. This process is further enhanced by our Temporal Guidance Sampling (TGS) technique, which encourages temporal consistency and human-object contact during inference.
Specifically, we estimate the object pose sequence over F frames $\vec{p}^{1:F} \in \mathbb{R}^{F \times 9}$, conditioned on the corresponding body motion sequence $\vec{\theta}^{1:F} \in \mathbb{R}^{F \times 21 \times 3}$.
Instead of relying solely on the estimated score from the model, we introduce additional guidance term into the Probability Flow ODE~\cite{score_based_modeling} as follows:
\begin{equation}
    \begin{gathered}
    \frac{d\vec{p}^{1:F}}{dt} = -\sigma(t)\dot{\sigma}(t) \Psi (\vec{p}^{1:F}, t | \vec{\theta^}{1:F})  + \nabla_{\vec{p}^{1:F}}{\mathcal{L_{\text{TGS}}}}.
    \end{gathered}
\end{equation}
The $\nabla_{\vec{p}^{1:F}}{\mathcal{L_{\text{TGS}}}}$ is the gradient of the object pose at time step $t$ on the following TGS loss, $\mathcal{L}_{\text{TGS}}$:
\begin{gather}
    \mathcal{L}_{\text{TGS}}(\vec{p}^{1:F}) = \lambda_{\text{s}}\mathcal{L}_{\text{smooth}} + \lambda_{\text{c}}\mathcal{L}_{\text{contact}},\\
    \mathcal{L}_{\text{smooth}} (\vec{p}^{1:F}) = \left\lVert \vec{p}^{2:F} - \vec{p}^{1:{F-1}} \right\rVert_2^2, \\
    \mathcal{L}_{\text{contact}} (\vec{p}^{1:F}) = \mathcal{D}_{\text{chamfer}}\left(\mathcal{V}_{h}, \mathcal{V}_{o} (\vec{p}^{1:F}) \right), 
\end{gather}
where $\mathcal{L}_{\text{smooth}}$ encourages temporal consistency between consecutive frames. The $\mathcal{L}_{\text{contact}}$ encourages proximity between human and object vertices, measured by the Chamfer Distance $\mathcal{D}_{\text{chamfer}}$ between the human vertices $\mathcal{V}_{h}$ and object vertices $\mathcal{V}_{o}$ positioned by the current pose $\vec{p}^{1:F}$. In practice, we use a certain threshold to focus on nearby points only when computing the Chamfer Distance. 
We utilize RK45 ODE solver~\cite{runge_kutta} to solve the ODE trajectory.

\section{Experiments}
\label{sec:experiments}

In this section, we conduct experiments to evaluate our method. 
In Section~\ref{subsec:qual}, we show 4D HOI motions generated by DAViD for various objects, demonstrating the efficacy of DAViD in learning dynamic HOI patterns.
In Section~\ref{subsec:quant}, we conduct a quantitative comparison with other baselines and a quantitative ablation study on our design choices.

\subsection{Datasets}
\label{subsec:datasets}

For qualitative evaluation, we consider 30 input 3D object meshes from various sources, obtained from BEHAVE~\cite{BEHAVE}, InterCap~\cite{InterCap}, ShapeNet~\cite{ShapeNet, PartNet}, SAPIEN~\cite{SAPIEN}, FullBodyManipulation Dataset~\cite{OMOMO}, and SketchFab~\cite{SketchFab}.
For quantitative evaluation, we use FullBodyManipulation Dataset~\cite{OMOMO}, which consists of the motion of both humans and objects.

\subsection{Baselines and Metric}
\label{subsec:baslines_metric}

Even though our goal is to learn dynamic HOI patterns for various objects, which is slightly different from typical text-driven HOI motion synthesis~\cite{CGHOI, HOIDiff, InterDreamer, TesMo, InterDiff, THOR}, we conduct quantitative comparisons on existing datasets for evaluation.
Specifically, we compare DAViD against OMOMO~\cite{OMOMO} and CHOIS~\cite{CHOIS} on FullBodyManipulation Dataset~\cite{OMOMO}, which provide available code and evaluate the quality of HOI, unlike other stuides~\cite{CGHOI, HOIDiff, THOR} that focus on evaluating human quality solely.
However, OMOMO~\cite{OMOMO} and CHOIS~\cite{CHOIS} generate human motion conditioned on object motion (or waypoints), making a direct comparison with ours challenging.
To address this, we compare the HOI motion generated by baselines using object motion (or waypoints) condition with the HOI motion generated by DAViD using human motion condition. Each condition is extracted from the test dataset of FullBodyManipulation~\cite{OMOMO}, enabling a comparison with the ground truth. Due to the different conditions, we compare only the quality of HOI for the fairness, specifically the precision, recall, and F1 score of the contact, following baselines~\cite{OMOMO, CHOIS}.
As DAViD performs modeling of object poses during interaction, we evaluate by cropping each test dataset sequence from the start of the interaction (contact start) to the end (contact end).
Additionally, to assess the human motion quality, we report the Foot Sliding (FS) and foot height (in centimeters) of the human motion generated from DAViD and the baselines, both trained on FullBodyManipulation Dataset~\cite{OMOMO}.

\subsection{Qualitative Results}
\label{subsec:qual}

\noindent\textbf{4D HOI Generation for Various Objects. }
Different from previous studies, we leverage the pre-trained video diffusion model, which allows to learn dynamic patterns of HOI for unbounded object categories.
Fig.~\ref{fig:qual} visualizes the HOI motion for various objects generated by DAViD.
The frames are visualized in temporal order. 
The results show that our method effectively models the dynamic patterns in HOIs.
In Fig.~\ref{fig:qual}~(a), Fig.~\ref{fig:qual}~(b), Fig.~\ref{fig:qual}~(c) we observe that the human motion and the relative position of the object are well modeled in relatively static scenarios.
In cases with more dynamic human motion, we find that various interaction patterns, including pulling$-$Fig.~\ref{fig:qual}~(d), pushing$-$Fig.~\ref{fig:qual}~(h), lifting$-$Fig.~\ref{fig:qual}~(g), and holding$-$Fig.~\ref{fig:qual}~(f) are well aligned with the typical use of objects of humans.
Also, we can observe that our method can effectively capture complex human motions and their corresponding object motions, such as back and forth movement, as shown in Fig.~\ref{fig:qual}~(e) and Fig.~\ref{fig:qual}~(j).
Importantly, unlike previous studies where motion translation typically resulted from walking, we observe that our model effectively captures cases of translation changes without foot movement, as shown in Fig.~\ref{fig:qual}~(i).

\begin{figure}[t]
\centering
\vspace{-2pt}
\includegraphics[width=\columnwidth]{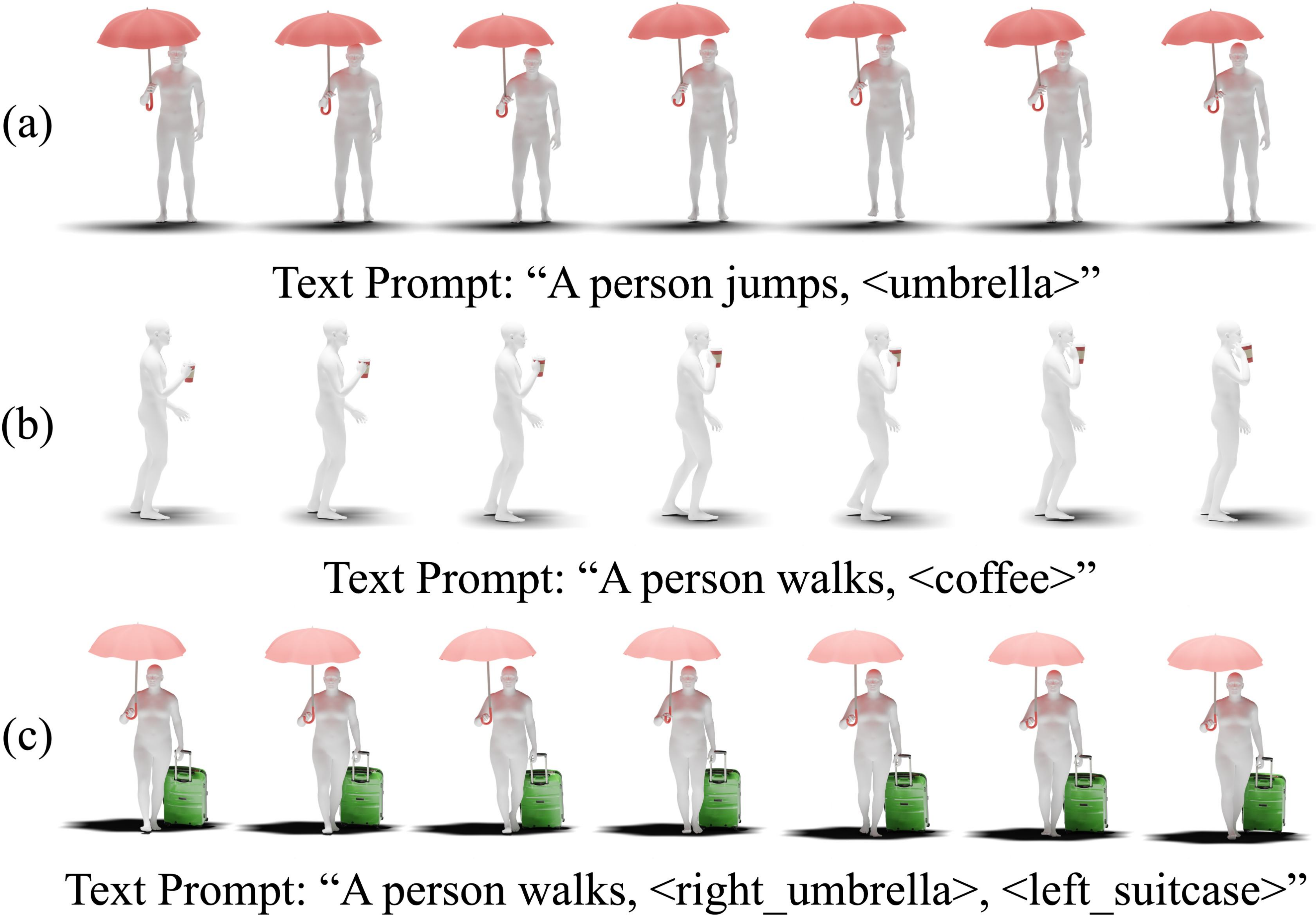}
\caption{\textbf{Combining Concepts.} 
Our HOI motion concepts can be combined with original knowledge of pre-trained MDM, even for multiple HOI motion concepts.
} 
\vspace{-1pt}
\label{fig:concepts}
\end{figure}

\begin{figure}[t]
\centering
\vspace{-5pt}
\includegraphics[width=\columnwidth]{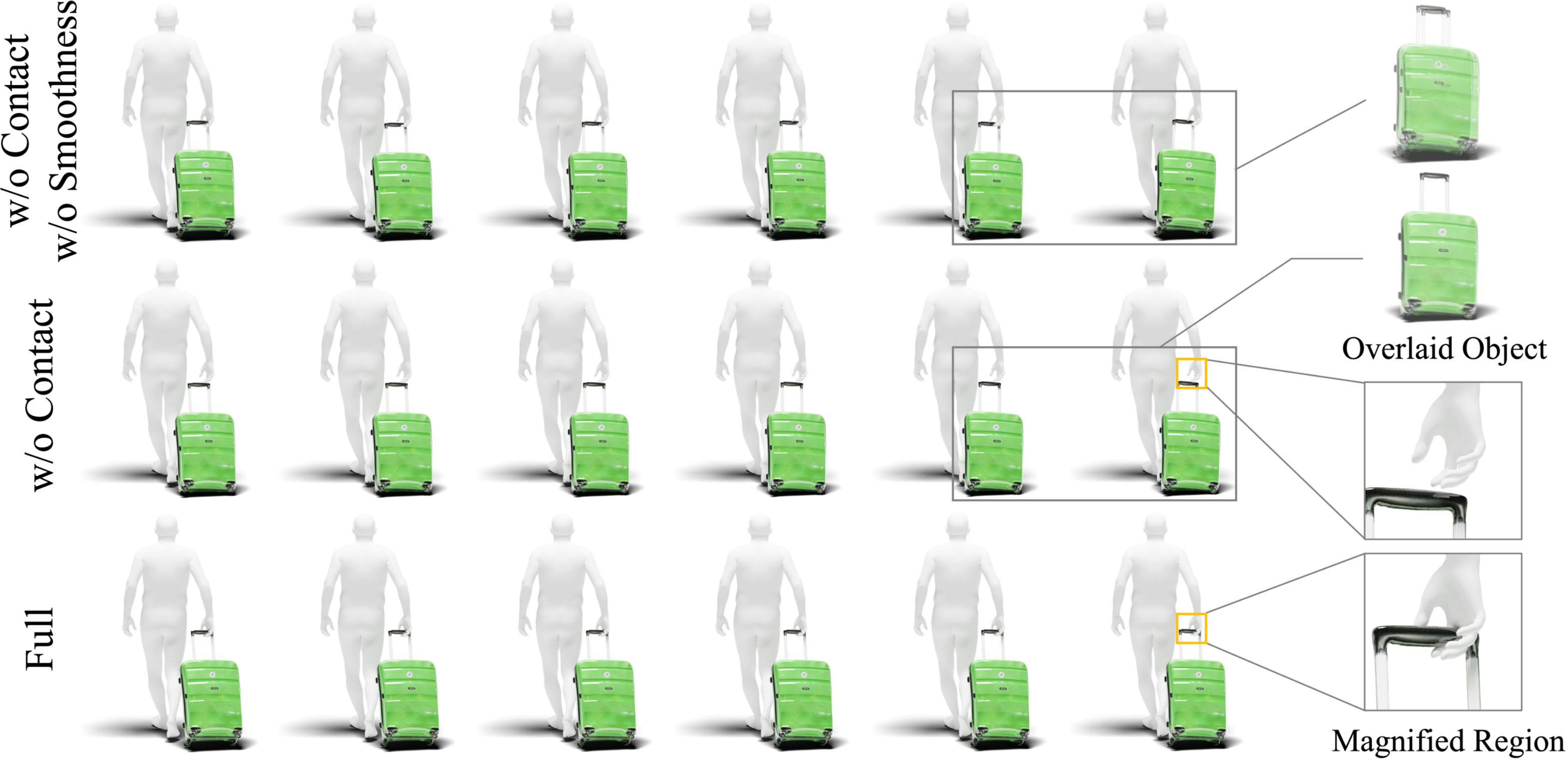}
\caption{\textbf{Ablation on Temporal Guidance Sampling.} 
Our guidance during the inference stage reduces object jitter and non-contact interactions, enabling the generation of plausible HOI motion.
} 
\label{fig:ablation_SGS}
\vspace{-10pt}
\end{figure}

\noindent\textbf{Combining Multiple HOI Concepts. }
As we model dynamic patterns of human motion using LoRA~\cite{LoRA} on a pre-trained MDM~\cite{MDM}, the existing knowledge of MDM~\cite{MDM} can be combined with the newly learned concepts of HOI.
Fig.~\ref{fig:concepts}~(a) and Fig.~\ref{fig:concepts}~(b) show the novel HOI motion generated by combining the pre-trained MDM's previously known jumping and walking concepts with the newly learned HOI concept; umbrella, and coffee.
By simply adding LoRA tags along with the text prompt, we can combine the newly learned HOI concept with the existing motions known by pre-trained MDM to create a novel human motion. The corresponding object motion is then generated based on the human motion, forming the novel HOI motion.

This approach goes beyond combining a single HOI motion concept as shown in Fig.~\ref{fig:concepts}~(c), and allows to combine multiple HOI motion concepts which share a similar body structure.
We merge two separately trained LoRA (suitcase, umbrella) parameters into the original network by a weighted summation, combining the two concepts into a single network. The object motions are then generated similarly based on the human motion.

\noindent\textbf{Ablation on Temporal Guidance Sampling. }
To verify the efficacy of our TGS, we conduct an ablation study.
Fig.~\ref{fig:ablation_SGS} shows six continuous frames of HOI motion with a suitcase, along with an overlay of objects in the subsequent two frames and a magnified view of the hand-object interaction.
Without smoothness guidance, object poses are independently sampled for the human pose, causing motion jitters between adjacent frames and reducing the plausibility of the motion.
Additionally, without contact guidance, fine details such as hand contact are lacking, resulting in floating objects.
This shows the importance of TGS as it finds a path that continuously connects plausible object poses for each frame, creating a motion that is frame-wise plausible, temporally smooth, and contacted during interaction.

\begin{table}[t]
    \small
    \centering
    \resizebox{1.0\columnwidth}{!}{
        \begin{tabular}{lccccc}
        \toprule
        Methods & $\text{C}_{\text{prec}} \uparrow$ & $\text{C}_{\text{rec}} \uparrow$ & $\text{F1 Score} \uparrow$ & $\text{FS} \downarrow$ & $\text{H}_{\text{feet}} \downarrow$ \\
        \midrule \midrule
        CHOIS~\cite{CHOIS} & 0.761 & 0.449 & 0.565 & 0.472 & 3.47\\
        $\text{DAViD}_{\text{w/o contact guidance}}$ & \textbf{0.923} & 0.154 & 0.264 & \multirow{2}{*}{\textbf{0.234}} & \multirow{2}{*}{\textbf{2.30}}  \\
        $\text{DAViD}_{\text{Full}}$ & 0.882 & \textbf{0.577} & \textbf{0.698} &  & \\
        \midrule
        OMOMO~\cite{OMOMO} & 0.783 & 0.547 & 0.644 & 0.255 & 2.58 \\
        $\text{DAViD}_{\text{w/o contact guidance}}$ & 0.800 & 0.186 & 0.302 & \multirow{2}{*}{\textbf{0.247}} & \multirow{2}{*}{\textbf{2.35}} \\
        $\text{DAViD}_{\text{Full}}$ & \textbf{0.828} & \textbf{0.558} & \textbf{0.667} &  &  \\
        \midrule
        MDM~\cite{MDM} & \text{N/A} & \text{N/A} & \text{N/A} & 0.260 & 2.48 \\
        \bottomrule
        \end{tabular}
    }
\caption{\textbf{Quantitative Results. } We evaluate the quality of generated HOI motion against FullBodyManipulation Dataset.}
\vspace{-5pt}
\label{tab:quant}
\end{table}

\begin{table}[t]
    \resizebox{0.4845\columnwidth}{!}{
        \begin{tabular}{lc}
        \toprule
        Methods & $\text{Jitter} \rightarrow$  \\
        \midrule \midrule
        $\text{DAViD}$ & \textbf{30.9}   \\
        $\text{DAViD}_{\text{w/o smoothness guidance}}$ & $2.95 \times 10^3$ \\
        \midrule
        $\text{FullBodyManip DB}$~\cite{OMOMO} & 10.7  \\
        \bottomrule
        \end{tabular}
    }
    \resizebox{0.5055\columnwidth}{!}{
        \begin{tabular}{lcc}
        \toprule
        Methods & $\text{FID} \downarrow$ & $\text{Diversity} \rightarrow$ \\
        \midrule \midrule
        $\text{DAViD}$ & \textbf{1.30} & \textbf{8.90}  \\
        $\text{DAViD}_{\text{w/o LoRA}}$ & 2.93 & 8.59 \\
        \midrule
        Real ~\cite{HumanML3D} & 0.00160 & 9.48 \\
        \bottomrule
        \end{tabular}
    }
\caption{\textbf{Quantitative Ablation.} Our smoothness guidance reduces the jitter level of the object motion (left), and our LoRA module better preserves the pre-trained MDM knowledge (right).}
\vspace{-10pt}
\label{tab:quant_ablation}
\end{table}

\subsection{Quantitative Results}
\label{subsec:quant}
\noindent\textbf{Comparison with Baselines. } To verify the efficacy of our model, we compare DAViD against OMOMO~\cite{OMOMO} and CHOIS~\cite{CHOIS} on the FullBodyManipulation Dataset~\cite{OMOMO}.
Specifically, we measure the performance of DAViD on each of the test datasets of the FullBodyManipulation~\cite{OMOMO}, selected from baselines.
For the metrics about HOI quality, hand contact is considered when the average distance between the 25 hand joints of SMPL-X~\cite{SMPLX} and the object vertices is smaller than a threshold, 0.05.
As shown in Tab.~\ref{tab:quant}, DAViD outperforms the baselines for all contact metrics.
This shows that DAViD generates plausible HOI motion, demonstrating the efficacy of our approach in modeling the object pose distribution from the human pose, and sampling object motion using TGS.
Additionally, we report FS and foot height of the human motions generated by DAViD, where we generate the same number of samples as in the test dataset.
We demonstrate that DAViD generates more natural human motions compared to the baselines, with lower FS and foot height, as shown in Tab.~\ref{tab:quant}.

\noindent\textbf{Ablation Study. } 
We conduct a quantitative ablation study on our key design choices, TGS and LoRA~\cite{LoRA}.
As shown in Tab.~\ref{tab:quant}, we find that our contact guidance in TGS is crucial for expressing detailed hand contact in HOI motions.
Note that this is not obvious, as contact guidance automatically detects potential contact region and guides them to get closer, and not provide information on where the contact should occur.
The result shows that the contact guidance in TGS performs fine refinement of object motion within a learned distribution.
We also perform an ablation study on our smoothness guidance in TGS by generating object motion using human motion from FullBodyManipulation~\cite{OMOMO} and measure the jitter.
We define jitter following MocapEvery~\cite{mocapevery} and compute it for object translation.
As shown in Tab.~\ref{tab:quant_ablation}~(left), our smoothness guidance in TGS significantly reduces the object jitter that occurs when sampling object poses frame-wise, bringing it to a level similar to the jitter found in real-world capture data.

We conduct an ablation study on LoRA to verify its ability to maintain the pre-trained model's knowledge while extending the model's space.
We train both models for 1000 epochs on the FullBodyManipulation dataset~\cite{OMOMO} and compare the FID and diversity on HumanML3D dataset~\cite{HumanML3D} (which MDM is trained on) for text-conditioned motion generation.
As shown in Tab.~\ref{tab:quant_ablation}~(right), LoRA enables the model to learn new concepts without significantly altering the motion distribution of the pre-trained model.
We demonstrate that our design, which adds LoRA to MDM is important to preserve the knowledge of pre-trained model, allowing to combine various HOI concepts with pre-trained human motion.

\section{Discussion}
\label{sec:discussion}

In this paper, we present DAViD, a method to learn Dynamic Affordance, addressing both static patterns (\eg, contact, spatial relation) and dynamic motion patterns in HOIs.
While obtaining a 4D HOI dataset including diverse objects is challenging, we observe that the video diffusion model has prior knowledge of the dynamic patterns for general objects.
Compared to learning affordance from videos generated from text-to-video models, our pipeline, which first renders the 3D object to generate HOI images and then uses an image-to-video model, has the advantage of lifting affordances in video into 3D space.
Using the 4D HOI samples generated through our lifting pipeline, we train DAViD, which consists of (1) the LoRA module of MDM for learning human motion patterns and (2) an object pose diffusion model with human pose condition.
We qualitatively show that the 4D HOI samples generated by DAViD effectively model the dynamic pattern of human pose and corresponding object pose during HOIs.
Quantitative results show that our DAViD models human motion and interactions that are more similar to real-world HOI than the baseline.
In addition, we demonstrate the advantage of using LoRA to model the dynamic patterns of using specific objects by generating HOI motion that includes multi-object interactions.

\section*{Acknowledgements}
This work was supported by NAVER Webtoon, NRF grant funded by the Korean government (MSIT) (No. RS-2022-NR070498, No. RS-2023-00218601, and RS-2025-25396144), and IITP grant funded by the Korea government (MSIT) (No. RS-2024-00439854, RS-2021-II211343, and RS-2025-25442338). H. Joo is the corresponding author.

{
    \small
    \bibliographystyle{ieeenat_fullname}
    \bibliography{main}
}

\clearpage
\setcounter{section}{0}
\setcounter{figure}{0}
\renewcommand*{\thesection}{\Alph{section}} 
\renewcommand{\theequation}{S.\arabic{equation}} 
\renewcommand{\thefigure}{S.\arabic{figure}} 
\renewcommand{\thealgorithm}{S.\arabic{algorithm}}

\section{Implementation Details}
\label{sec:implementation}

In this section, we introduce the details of our method for modeling Dynamic Affordance.
From Sec.~\ref{subsec:rendering} to Sec.~\ref{subsec:lifting}, we cover our first pipeline, 4D HOI Sample Generation. 
Sec.~\ref{subsec:network} and Sec.~\ref{subsec:training} describe our second pipeline, learning Dynamic Affordance.

\subsection{Rendering Object from Multi-Viewpoints}
\label{subsec:rendering}

For camera installation, we position eight perspective cameras evenly spaced at 45\degree intervals around the object at a fixed elevation of 5\degree.
The radius (distance of camera to origin) is set as a hyperparameter along with additional adjustment of camera's z-coordinate to ensure the object fits within the image frame.
To have a consistent camera setup in the uplifting pipeline, we follow GVHMR~\cite{gvhmr} and set the intrinsic parameters as follows.
\begin{equation}
    K = \begin{bmatrix}
        f & 0 & w/2 \\
        0 & f & h/2 \\
        0 & 0 & 1 
    \end{bmatrix},
\end{equation}
where $f = \sqrt{h^2 + w^2}$ and $h$, $w$ represent the height and width of our rendering image, respectively.
In practice, we use $h=800$, $w=1200$ for rendering.
For object installation, relatively large and stationary ground-placed objects (e.g., motorcycles) are placed at the origin in a canonical state, while small and portable objects (e.g., umbrellas) are perturbed by sampling their position and rotation within a certain range.
The range of the position and rotation is set as a hyperparameter.

\subsection{Generating 2D HOI images}
\label{subsec:gen_images}
For the image rendered in Sec.~\ref{subsec:rendering}, we use the Canny edge detector~\cite{canny} to obtain structural guidance.
In practice, we use an upper threshold of 30 and a lower threshold of 25 to capture dense structures.
We use the obtained Canny edges as input of ControlNet~\cite{ControlNet} and leverage the off-the-shelf pre-trained 2D diffusion model, FLUX~\cite{flux}, to generate the 2D HOI Image.
Unlike other approaches~\cite{ComA} that directly use inpainting on the rendered object, maintaining a consistent background color (e.g., white, gray), our method generate background, offering the advantage of aligning with the training domain of the video diffusion model while providing motion cues to the world-grounded HMR (\eg, if the background moves left, the subject moves right).
For specific settings, we use a classifier-free guidance scale of 3.5, 28 inference steps, and the FlowMatchEulerDiscrete scheduler~\cite{flowmatch} for image generation.
In cases where it is natural for a person to occlude an object (\eg, a hand occluding the handle of a cart), strong structural guidance can lead to the generation of implausible images.
Therefore, we set the ControlNet~\cite{ControlNet} conditioning guidance as 0.725 for the first 12 denoising steps, and 0.0 for the later steps.
We empirically find that this approach helps generating plausible HOI image considering appropriate occlusion.
For the text prompt for generating images, we use a vision-language model~\cite{gpt4o} to automatically obtain prompts that include HOI. 
Specifically, we obtain the text prompt using the following input.
\begin{quote}
    \textit{Write a text prompt in two sentence. The format of the text prompt should start with ``1 person'' and should include word ``\{category\}''. Write a detailed text prompt focusing on human pose and the interaction between ``1 person'' and ``\{category\}''. The third word of the first sentence must describe the interaction.}
\end{quote}
We add the additional tag ``, full body'' at the end of the obtained text prompt, which we find beneficial for expressing the holistic body in image.
While we know the category of the input 3D object in many cases, we use the rendering of the object to request a prompt if the category is not available.

\begin{figure*}[t]\centering
\includegraphics[width=\linewidth, trim={0 0 0 0},clip]{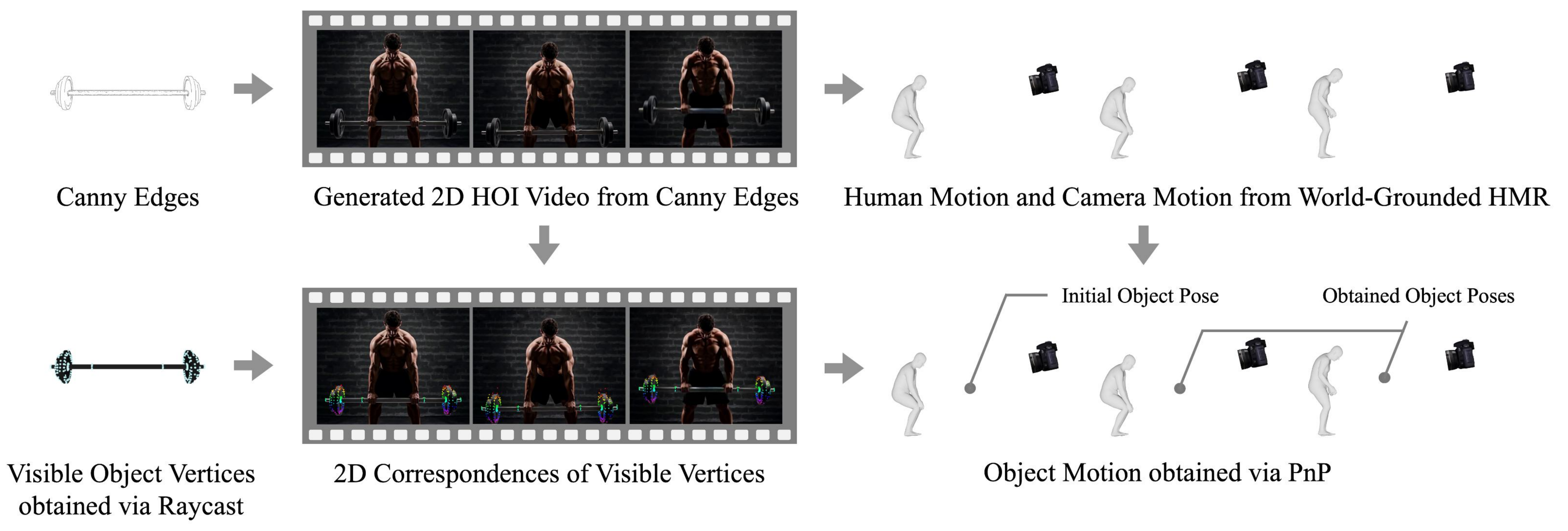}
\captionof{figure}{\textbf{Obtaining Object Motion.} We first leverage off-the-shelf world-grounded HMR to obtain human motion and corresponding camera motion. Then, for the object vertices visible in our rendering camera, we find the 2D correspondences across the video. Using the 2D-3D correspondence of the vertices and camera pose for every frame, we compute the object pose for each frame via PnP.
}
\label{fig:lifting}
\vspace{-10pt}
\end{figure*}

\begin{figure*}[t]\centering
\includegraphics[width=\linewidth, trim={0 0 0 0},clip]{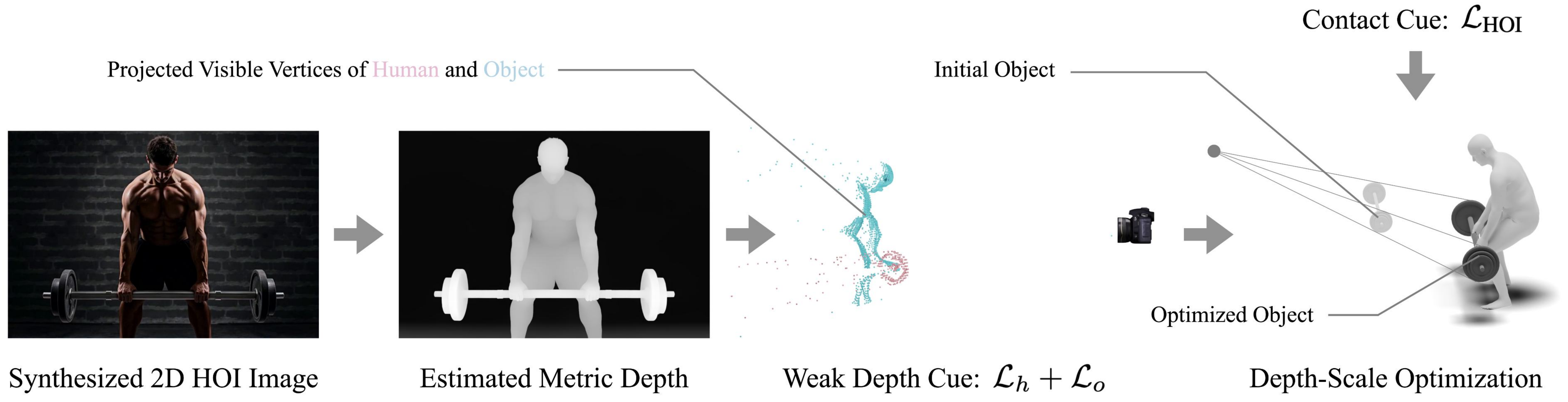}
\captionof{figure}{\textbf{Resolving Depth Ambiguity.} To resolve the depth ambiguity between human and object motion, we leverage weak depth cues obtained from a metric depth model and contact cues, based on the intuition that object movement is driven by human contact. By optimizing the human and object scales using these cues, we obtain the 4D HOI sample.
}
\label{fig:optimization}
\end{figure*}

\subsection{Generating 2D HOI Video from 2D HOI Image}
\label{subsec:video_gen}
We use a pre-trained video diffusion model~\cite{kling} to generate 2D HOI videos from 2D HOI images.
For the text prompt, we use the same one used for generating the 2D HOI image.
As the video diffusion model support only specific resolution conditions, we resize both the input image and the output video.

\subsection{Lifting 2D HOI Videos to 4D HOI Samples}
\label{subsec:lifting}
We detail the process of (1) computing object motion and (2) resolving depth ambiguity which are used to lift 2D HOI Videos into 4D HOI samples with additional figures (Fig.~\ref{fig:lifting}, Fig.~\ref{fig:optimization}).

\noindent\textbf{Obtaining Object Motion. } We leverage an off-the-shelf world-grounded HMR, GVHMR~\cite{gvhmr} to obtain both human motion and the corresponding camera motion in world coordinates.
The core idea for obtaining the remaining object motion is to find 2D-3D correspondences for each frame.
As we use a camera model same with GVHMR~\cite{gvhmr} for rendering, it is possible to transform (rotation and translation of) the rendering camera to the first frame camera of GVHMR's output.
Using the same transformation, we obtain the initial (first frame) object pose aligned with the human and camera motion.
At the same time, we obtain the vertices of the object visible in the rendered camera through raycasting~\cite{raycast}, and find the correspondences of 2D projection points across the generated 2D HOI Video via video tracking~\cite{cotracker, cotracker3}.
Through this, we establish the 2D-3D correspondences of the vertices for each frame with known camera motion, allows PnP~\cite{PnP, EPnP} to compute object pose for each frame, as shown in Fig.~\ref{fig:lifting}.

\noindent\textbf{Resolving Depth Ambiguity. } Even after obtaining the human motion, camera motion, and object motion aligned on 2D, the human motion and object motion do not interact with each other in 3D space.
To resolve the depth ambiguity that occurs on perspective camera rays, we optimize the object's scale in the first frame using (1) weak depth cues and (2) contact cues.
First, we use a publicly available depth estimation model~\cite{depth_pro} to predict the metric depth from the generated images.
As shown in Fig.~\ref{fig:optimization}, the visible vertices of the human and the object, obtained through raycasting~\cite{raycast} are projected into 3D space to construct a point cloud.
The MSE distance between the human point cloud and the corresponding visible 3D vertices of the human is defined as $\mathcal{L}_h$, and we define $\mathcal{L}_o$ similarly.
Additionally, based on the intuition that the object must be in contact with the human to have movement, we define $\mathcal{L}_{\text{HOI}}$ as the loss, calculated as the average distance of the $n$ closest 3D vertices of the object to the 3D vertices of the human. 
In practice, we set $n$ to one-third of the total number of vertices in the object mesh.
We define the final loss as $\mathcal{L}_{\text{total}} = \mathcal{L}_h + \mathcal{L}_o + \mathcal{L}_{\text{HOI}}$ and optimize the scales of the human and object, $s_h$, and $s_o$, to obtain to $s_h^*$, and $s_o^*$.
To preserve the real-world scale of the human, we fix the human scale and only adjust the object's scale by $s_o^* / s_h^*$.

\begin{figure*}[t]\centering
\includegraphics[width=\linewidth, trim={0 0 0 0},clip]{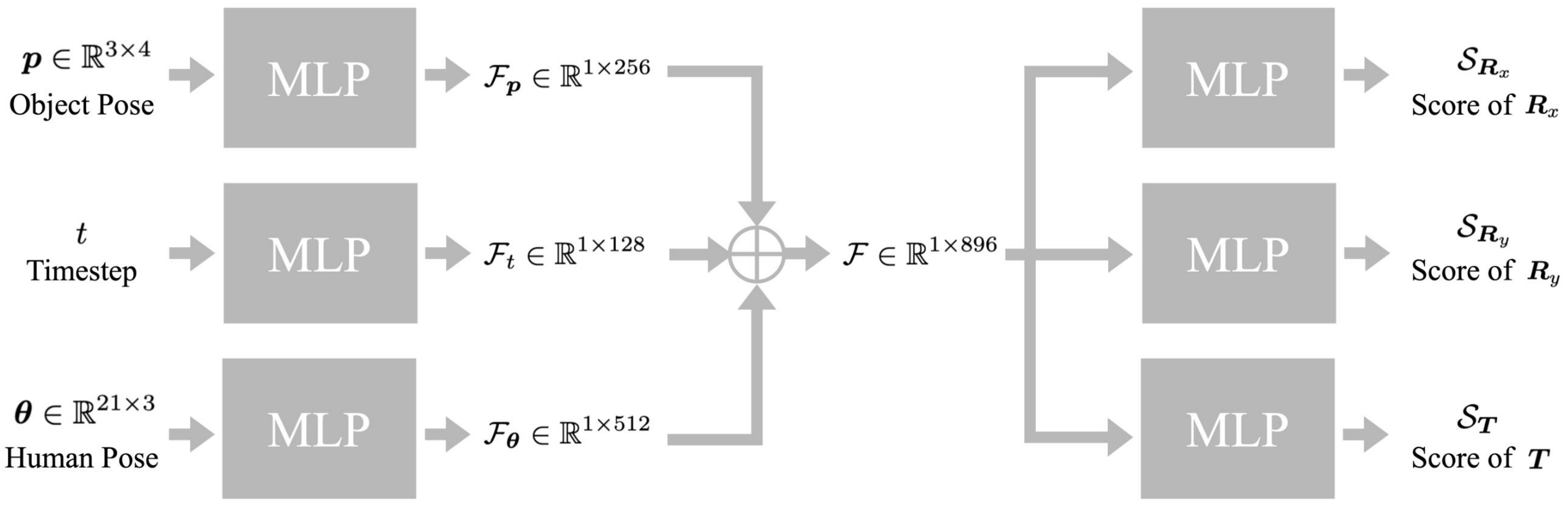}
\captionof{figure}{\textbf{Architecture of Human Conditioned Object Pose Diffusion Model.} We design a diffusion model that generates a plausible object pose for interacting with a given human pose. Each object pose, time step, and human pose are encoded by MLP. The concatenated features then pass through different MLPs, producing an object pose output consisting of 6D rotation and translation.
}
\label{fig:architecture}
\end{figure*}

\subsection{Network Architecture}
\label{subsec:network}
We describe the network architecture of (1) LoRA for MDM and (2) the Human-Conditioned Object Pose Diffusion Model, which form our DAViD.

\noindent\textbf{LoRA for MDM. } 
To learn concepts through LoRA~\cite{LoRA}, we model the concepts represented by the samples using text prompts.
To ensure that the text effectively models the concepts demonstrated by the given samples, we add LoRA~\cite{LoRA} layers to the multi-head attention within the transformer encoder layer of the pre-trained MDM.
Specifically, we add four 2-layer MLPs for query, key, value, and output projection, respectively for a single transformer encoder layer, allocating them as a space to learn additional knowledge.
We add this to all 8 transformer encoder layers stacked in the transformer encoder.

\noindent\textbf{Human Conditioned Object Pose Diffusion Model. } 
To model the conditional object pose based on the given human pose, we design a score-based diffusion model.
We encode the object pose, timestep, and human pose using each MLP, concatenating the feature vectors to construct the total feature.
The feature is then fed into three different MLPs, which output the scores for $\vec{R}_x$, $\vec{R}_y$, and $\vec{T}$, where $\vec{R}_x$ and $\vec{R}_y$ constitute the 6D rotation representation, and $\vec{T}$ represents the translation. 
The overall architecture is shown in Fig.~\ref{fig:architecture}.

\subsection{Training Details}
\label{subsec:training}
In this section, we describe the training details of (1) LoRA for MDM and (2) the Human Conditioned Object Pose Diffusion Model, which form our DAViD.

\noindent\textbf{LoRA for MDM. } 
For training the LoRA~\cite{LoRA} layer in the pre-trained MDM~\cite{MDM}, we create a dataset by extracting only the human motion from previously generated 4D HOI samples and processing it following HumanML3D~\cite{HumanML3D}.
The number of training samples varies by object category, ranging from $5$ to $50$, and we figure out that this amount is sufficient for learning the concept of human motion through LoRA~\cite{LoRA}.
During training, we freeze all other weights and train only the weights of the LoRA~\cite{LoRA} layer.
As our concepts are represented in the form of text, we use object category as a text prompt for training our LoRA.
For motions with multiple modes (e.g., left and right hand-object interactions), the text prompt is modified by adding tags such as ``left\_" or ``right\_'' before the main tag. We found that simply adding these additional tags gives controllability to model. We train a total of 500 to 3000 steps (depending on categories) using the Adam~\cite{adam} optimizer with a learning rate of $1\times 10^{-4}$ without decay.

\noindent\textbf{Human Conditioned Object Pose Diffusion Model. } 
For training human conditioned object pose diffusion model, we extract pairwise human pose and object pose from each frame of the 4D HOI Sample and use them as training data.
The number of data samples used varies by object category, ranging from $765$ to $7,650$.
We train total of 1000 to 5000 steps (depending on categories) using the Adam~\cite{adam} optimizer with a learning rate of $5\times 10^{-3}$ and a weight decay of 0.99.

\section{Experimental Details}
\label{sec:experimental_details}

\subsection{Additional Qualitative Results}
\label{subsec:add_qual}
\begin{figure*}[t]\centering
\includegraphics[width=\linewidth, trim={0 0 0 0},clip]{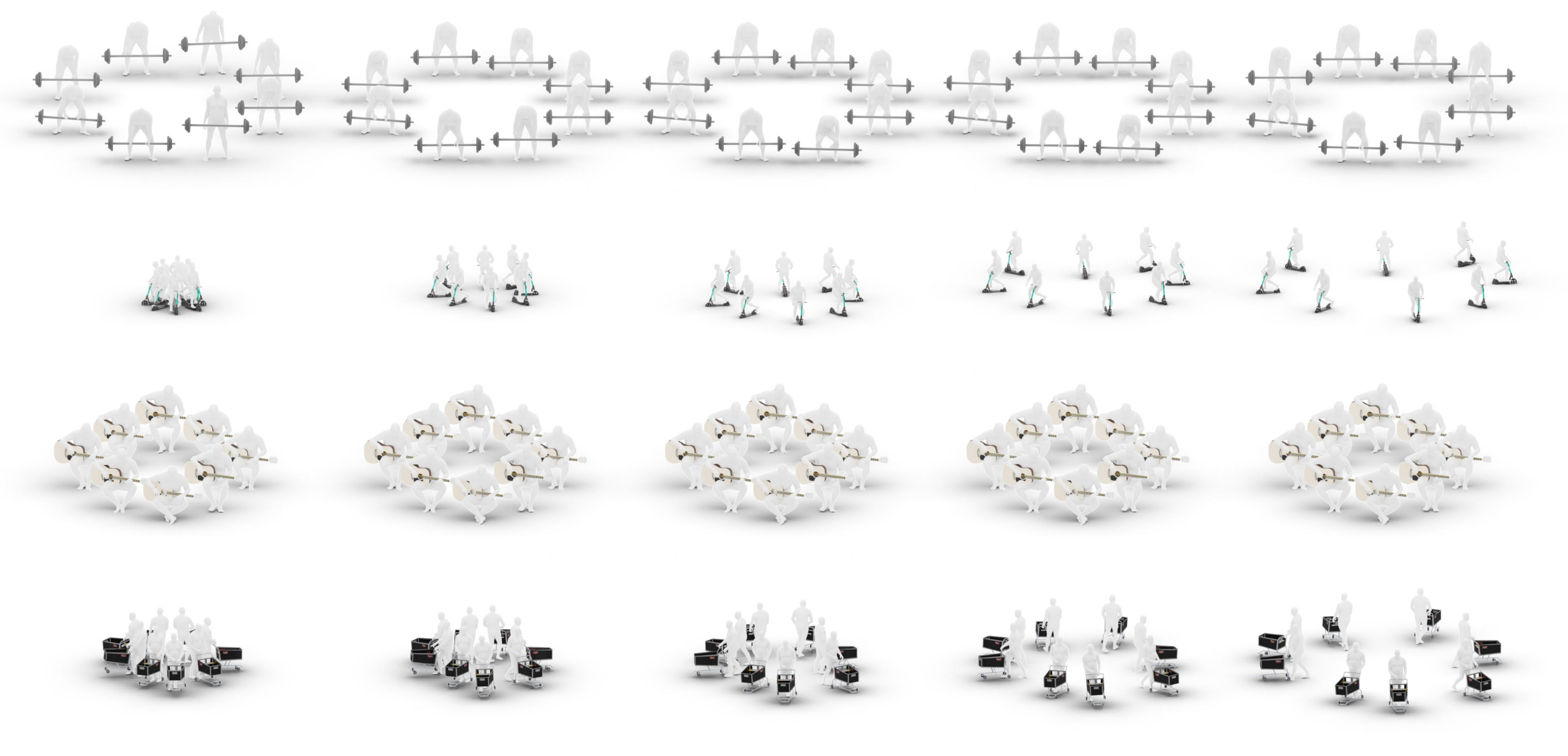}
\captionof{figure}{\textbf{Additional Qualitative Results.}
We showcase additional results of our method. We present diverse samples generated from our DAViD, with each frame visualized in temporal order.
}
\label{fig:addqual}
\end{figure*}

\begin{figure}[t]
\centering
\includegraphics[width=\columnwidth]{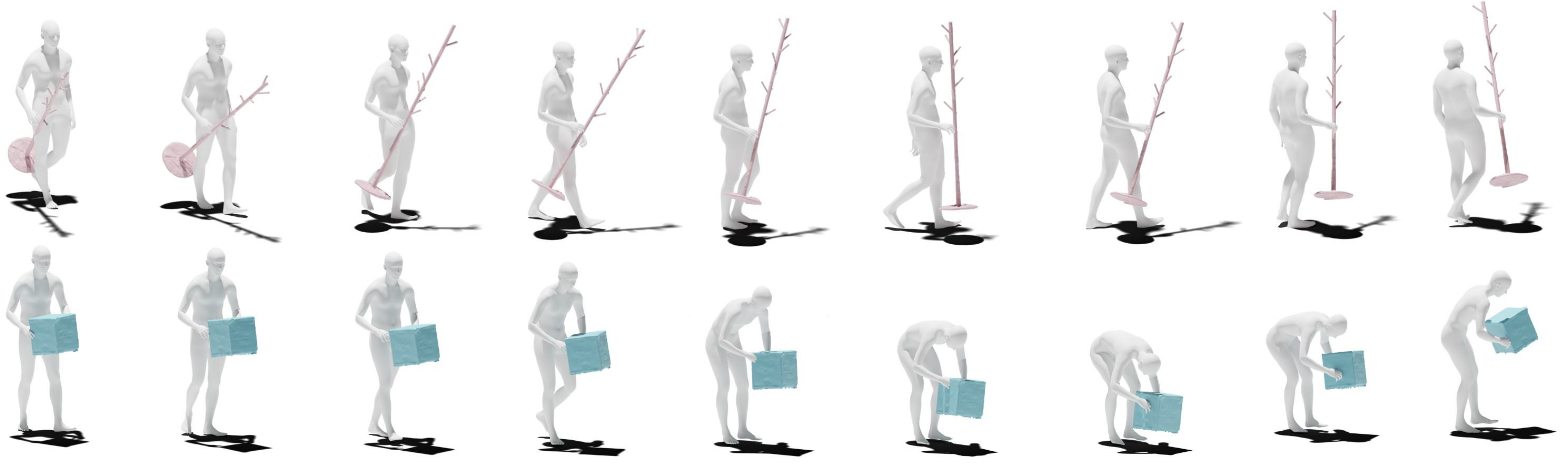}
\caption{\textbf{Qualitative Results on FullBodyManip Datatset.} We showcase additional qualitative results of DAViD trained on the FullBodyManip dataset.} 
\label{fig:david_fullbodymanip}
\vspace{-15pt}
\end{figure}

We showcase additional qualitative results in Fig.~\ref{fig:addqual} and Fig.~\ref{fig:david_fullbodymanip}.
In Fig.~\ref{fig:addqual}, we show the results of generating various HOI motions using our trained DAViD.
Through the results of generating various HOI motions, we demonstrate that our LoRA~\cite{LoRA} faithfully learns the dynamic patterns during HOI.
In Fig.~\ref{fig:david_fullbodymanip}, we show the qualitative results generated from DAViD, trained on the FullBodyManip~\cite{OMOMO} dataset.
We demonstrate that DAViD is not only able to learn coherent and simple HOI patterns, but also capable of generating relatively complex HOI motions.

\subsection{Additional Quantitative Results}
\label{subsec:add_quant}
We report additional comparisons with our baselines for each category of the FullBodyManip dataset in Tab.~\ref{tab:add_quant}.
As our TGS automatically detects potential contact points and guides them closer during sampling, we vary the threshold of potential contact $\rho$ to examine the effect of our guidance.
Note that the potential contact threshold $\rho$ used in TGS is lower than the threshold used for evaluating the metric (0.05), to ensure that points not considered as contact by the model are not forced into contact.
As shown in Tab.~\ref{tab:add_quant}, our contact guidance in TGS significantly improves recall and consequently the F1 score. We demonstrate that our contact guidance allows to sample fine-grained hand-object contact on the coarse distribution learned by our object pose diffusion model.
In contrast, precision tends to remain stable or decrease as the threshold level increases, which appears to be a side effect of unintended potential contacts detected in the early stage of the denoising preocess.
Empirically, we find that maintaining a low threshold around 0.02 minimizes side effects and is effective for sampling fine-grained hand-object contact.

\begin{table*}[t]
    \small
    \centering
    \resizebox{1.0\textwidth}{!}{
        \begin{tabular}{lcccccccccccccccccccccccccccc}
        \toprule
        \toprule
        
        \multirow{2}{*}{Methods} & \multicolumn{3}{c}{\text{Clothesstand}}  & \multicolumn{3}{c}{\text{Floorlamp}} & \multicolumn{3}{c}{\text{Largebox}} & \multicolumn{3}{c}{\text{Largetable}} & \multicolumn{3}{c}{\text{Monitor}} & \multicolumn{3}{c}{\text{Plasticbox}} & \multicolumn{3}{c}{\text{Smallbox}}  \\
        \cmidrule(lr){2-4} \cmidrule(lr){5-7} \cmidrule(lr){8-10} \cmidrule(lr){11-13} \cmidrule(lr){14-16} \cmidrule(lr){17-19} \cmidrule(lr){20-22}
         & $\text{C}_{\text{prec}} \uparrow$ & $\text{C}_{\text{rec}} \uparrow$ & $\text{F1} \uparrow$ & $\text{C}_{\text{prec}} \uparrow$ & $\text{C}_{\text{rec}} \uparrow$ & $\text{F1} \uparrow$ & $\text{C}_{\text{prec}} \uparrow$ & $\text{C}_{\text{rec}} \uparrow$ & $\text{F1} \uparrow$ & $\text{C}_{\text{prec}} \uparrow$ & $\text{C}_{\text{rec}} \uparrow$ & $\text{F1} \uparrow$ & $\text{C}_{\text{prec}} \uparrow$ & $\text{C}_{\text{rec}} \uparrow$ & $\text{F1} \uparrow$ & $\text{C}_{\text{prec}} \uparrow$ & $\text{C}_{\text{rec}} \uparrow$ & $\text{F1} \uparrow$ & $\text{C}_{\text{prec}} \uparrow$ & $\text{C}_{\text{rec}} \uparrow$ & $\text{F1} \uparrow$ \\
        \midrule \midrule
        $\text{DAViD}_{\rho=0.00}$ & 0.667 & 0.088 & 0.156 & 0.611 & 0.297 & 0.400 & 0.894 & 0.578 & 0.702 & 0.917 & 0.500 & 0.647 & 0.807 & 0.38 & 0.517 & 0.934 & 0.463 & 0.619 & 0.988 & 0.464 & 0.632 \\
        $\text{DAViD}_{\rho=0.01}$ & 0.913 & 0.309 & 0.462 & 0.500 & 0.486 & 0.493 & 0.862 & 0.311 & 0.457 & 0.743 & 0.197 & 0.311 & 0.847 & 0.485 & 0.617 & 0.848 & 0.275 & 0.415 & 0.985 & 0.515 & 0.676 \\
        $\text{DAViD}_{\rho=0.02}$  & 0.765 & 0.382 & 0.510 & 0.531 & 0.459 & 0.493 & 0.889 & 0.497 & 0.637 & 0.929 & 0.598 & 0.728 & 0.799 & 0.442 & 0.570 & 0.865 & 0.553 & 0.675 & 0.986 & 0.508 & 0.671 \\
        $\text{DAViD}_{\rho=0.03}$  & 0.536 & 0.221 & 0.313 & 0.600 & 0.243 & 0.346 & 0.863 & 0.547 & 0.669 & 0.837 & 0.545 & 0.661 & 0.823 & 0.523 & 0.640 & 0.821 & 0.471 & 0.599 & 0.982 & 0.545 & 0.701 \\
        CHOIS & 0.615 & 0.353 & 0.449 & 0.667 & 0.378 & 0.483 & 0.773 & 0.211 & 0.332 & 0.783 & 0.136 & 0.232 & 0.827 & 0.156 & 0.263 & 0.674 & 0.119 & 0.202 & 0.957 & 0.239 & 0.382 \\
        \midrule
        $\text{DAViD}_{\rho=0.00}$ & 0.667 & 0.098 & 0.171 & 0.784 & 0.138 & 0.235 & 0.951 & 0.485 & 0.642 & 0.808 & 0.371 & 0.509 & 0.849 & 0.345 & 0.491 & 0.881 & 0.349 & 0.500 & 0.991 & 0.528 & 0.689 \\
        $\text{DAViD}_{\rho=0.01}$ & 0.429 & 0.197 & 0.270 & 0.615 & 0.114 & 0.193 & 0.979 & 0.539 & 0.695 & 0.839 & 0.173 & 0.287 & 0.895 & 0.488 & 0.632 & 0.824 & 0.371 & 0.512 & 0.990 & 0.610 & 0.755 \\
        $\text{DAViD}_{\rho=0.02}$ & 0.405 & 0.279 & 0.330 & 0.667 & 0.276 & 0.391 & 0.954 & 0.488 & 0.645 & 0.873 & 0.456 & 0.599 & 0.911 & 0.602 & 0.725 & 0.800 & 0.446 & 0.572 & 0.989 & 0.649 & 0.784  \\
        $\text{DAViD}_{\rho=0.03}$ & 0.571 & 0.328 & 0.417 & 0.472 & 0.119 & 0.190 & 0.959 & 0.588 & 0.729 & 0.868 & 0.533 & 0.661 & 0.877 & 0.564 & 0.687 & 0.790 & 0.522 & 0.629 & 0.990 & 0.645 & 0.781 \\
        OMOMO & 0.432 & 0.311 & 0.362 & 0.828 & 0.114 & 0.201 & 0.917 & 0.617 & 0.738 & 0.868 & 0.603 & 0.711 & 0.776 & 0.432 & 0.555 & 0.831 & 0.342 & 0.484 & 0.982 & 0.639 & 0.774 \\
        \midrule
        \midrule
        
        \multirow{2}{*}{Methods} & \multicolumn{3}{c}{\text{Smalltable}}  & \multicolumn{3}{c}{\text{Suitcase}} & \multicolumn{3}{c}{\text{Trashcan}} & \multicolumn{3}{c}{\text{Tripod}} & \multicolumn{3}{c}{\text{Whitechair}} & \multicolumn{3}{c}{\text{Woodchair}} & \multicolumn{3}{c}{\text{Average}}  \\
        \cmidrule(lr){2-4} \cmidrule(lr){5-7} \cmidrule(lr){8-10} \cmidrule(lr){11-13} \cmidrule(lr){14-16} \cmidrule(lr){17-19} \cmidrule(lr){20-22}
         & $\text{C}_{\text{prec}} \uparrow$ & $\text{C}_{\text{rec}} \uparrow$ & $\text{F1} \uparrow$ & $\text{C}_{\text{prec}} \uparrow$ & $\text{C}_{\text{rec}} \uparrow$ & $\text{F1} \uparrow$ & $\text{C}_{\text{prec}} \uparrow$ & $\text{C}_{\text{rec}} \uparrow$ & $\text{F1} \uparrow$ & $\text{C}_{\text{prec}} \uparrow$ & $\text{C}_{\text{rec}} \uparrow$ & $\text{F1} \uparrow$ & $\text{C}_{\text{prec}} \uparrow$ & $\text{C}_{\text{rec}} \uparrow$ & $\text{F1} \uparrow$ & $\text{C}_{\text{prec}} \uparrow$ & $\text{C}_{\text{rec}} \uparrow$ & $\text{F1} \uparrow$ & $\text{C}_{\text{prec}} \uparrow$ & $\text{C}_{\text{rec}} \uparrow$ & $\text{F1} \uparrow$ \\
        \midrule \midrule

        $\text{DAViD}_{\rho=0.00}$  & 0.926 & 0.482 & 0.634 & 0.990 & 0.564 & 0.718 & 0.961 & 0.508 & 0.665 & 0.714 & 0.195 & 0.306 & 0.706 & 0.213 & 0.328 & 0.913 & 0.328 & 0.483 & 0.848 & 0.389 & 0.524 \\
        $\text{DAViD}_{\rho=0.01}$  & 0.953 & 0.527 & 0.679 & 0.992 & 0.609 & 0.755 & 0.952 & 0.496 & 0.652 & 0.727 & 0.312 & 0.436 & 0.808 & 0.187 & 0.303 & 0.716 & 0.276 & 0.398 & 0.847 & 0.383 & 0.511\\
        $\text{DAViD}_{\rho=0.02}$  & 0.915 & 0.691 & 0.788 & 0.979 & 0.609 & 0.751 & 0.916 & 0.628 & 0.745 & 0.731 & 0.247 & 0.369 & 0.825 & 0.609 & 0.701 & 0.779 & 0.495 & 0.605 & \textbf{0.850} & \textbf{0.517} & \textbf{0.634}\\
        $\text{DAViD}_{\rho=0.03}$  & 0.907 & 0.717 & 0.801 & 0.967 & 0.649 & 0.777 & 0.947 & 0.893 & 0.919 & 0.667 & 0.208 & 0.317 & 0.826 & 0.591 & 0.689 & 0.711 & 0.448 & 0.55 & 0.812 & 0.508 & 0.614 \\
        CHOIS & 0.944 & 0.431 & 0.592 & 0.942 & 0.251 & 0.397 & 0.580 & 0.165 & 0.257 & 0.917 & 0.143 & 0.247 & 0.483 & 0.062 & 0.110 & 0.717 & 0.224 & 0.341 & 0.760 & 0.221 & 0.330 \\
        \midrule
        $\text{DAViD}_{\rho=0.00}$ & 0.936 & 0.457 & 0.614 & 0.989 & 0.420 & 0.590 & 0.970 & 0.561 & 0.711 & 0.839 & 0.173 & 0.287 & 1.00 & 0.255 & 0.407 & 0.922 & 0.371 & 0.529 & \textbf{0.891} & 0.350 & 0.490 \\
        $\text{DAViD}_{\rho=0.01}$ & 0.931 & 0.502 & 0.653 & 0.986 & 0.501 & 0.664 & 0.963 & 0.630 & 0.761 & 0.902 & 0.306 & 0.457 & 0.963 & 0.265 & 0.416 & 0.810 & 0.406 & 0.540 & 0.856 & 0.392 & 0.526 \\
        $\text{DAViD}_{\rho=0.02}$ & 0.920 & 0.639 & 0.755 & 0.976 & 0.536 & 0.692 & 0.957 & 0.760 & 0.847 & 0.949 & 0.347 & 0.508 & 1.00 & 0.378 & 0.548 & 0.865 & 0.490 & 0.618 & 0.867 & 0.488 & 0.616 \\
        $\text{DAViD}_{\rho=0.03}$ & 0.945 & 0.706 & 0.809 & 0.985 & 0.552 & 0.708 & 0.956 & 0.699 & 0.807 & 0.967 & 0.321 & 0.482 & 0.974 & 0.388 & 0.555 & 0.774 & 0.516 & 0.619 & 0.856 & \textbf{0.499} & \textbf{0.621} \\
        OMOMO & 0.870 & 0.470 & 0.610 & 0.974 & 0.660 & 0.787 & 0.739 & 0.500 & 0.596 & 0.824 & 0.225 & 0.354 & 0.967 & 0.296 & 0.453 & 0.779 & 0.433 & 0.557 & 0.830 & 0.434 & 0.552 \\
        \bottomrule
        \bottomrule
        \end{tabular}
    }
\caption{\textbf{Additional Quantitative Results.} We report additional quantitative results for each category of the FullBodyManip dataset by varying the contact loss threshold used in our TGS.}
\vspace{-5pt}
\label{tab:add_quant}
\end{table*}

\subsection{Scale of the Object}
\label{subsec:scale}
As our human conditioned object pose diffusion model generates plausible object pose for a given human pose, it does not provide information about the object's scale.
Since the output human of MDM and the human in the training data both have a uniform scale of 1.0, we automatically determine the appropriate object scale in the generated HOI motion by sampling between the minimum and maximum scales of objects existing in our 4D HOI Samples.

\subsection{Generalizability Across Input 3D Objects.} 
\label{subsec:generalizability}
\begin{figure}[t]
\centering
\includegraphics[width=\columnwidth]{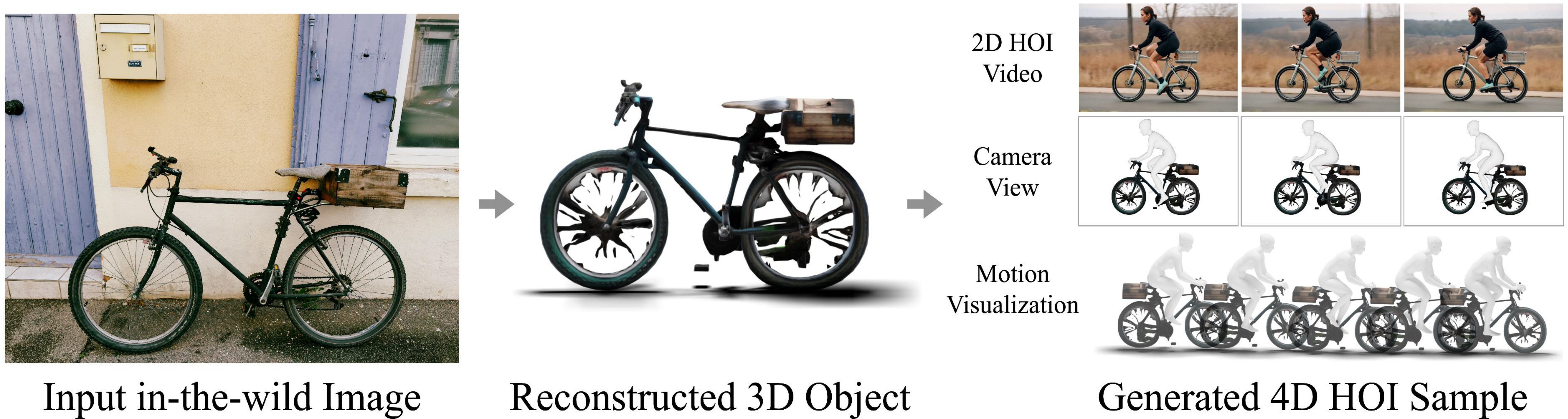}
\caption{\textbf{Generalizability Across Input 3D Objects.} Our 4D HOI sample generation pipeline is generalizable to any input 3D object, including those reconstructed from images.} 
\label{fig:inthewild}
\end{figure}

By leveraging pre-trained 2D diffusion models, our 4D HOI sample generation pipeline is scalable to various object categories and instances.
As shown in Fig.~\ref{fig:inthewild}, our pipeline allows to generate 4D HOI samples not only from 3D objects in existing datasets, but also from those reconstructed from in-the-wild images.
For the given in-the-wild image, we first reconstruct 3D objects from the image via TRELLIS~\cite{TRELLIS}, using them as input to our pipeline to generate the 4D HOI sample interacting with the object.

\section{Limitations and Future Work}
\label{sec:limitations_futurework}

\subsection{Spatial Bias on 2D HOI Image Generation}
\label{subsec:diversity}
Due to the internal spatial bias of the pre-trained 2D diffusion model, the model may fail to generate plausible images when structural guidance is introduced in locations that do not align with this bias, leading to collapse or hallucination.
For example, if an umbrella, which should be held by a hand, is rendered at the bottom of the image and Canny edges~\cite{canny} are extracted from it to generate an image, the model may create and use a new umbrella in a different location, rather than in the rendered region.
As a future direction, we can consider a new form of conditional image generation that is guided only by the structure of the given object, without guidance on its location in the image.
This approach is expected to remove the human labor we used for filtering malicious images.

\subsection{Limits of Smoothness Guidance Sampling}
\label{subsec:sgs}
Although our human-conditioned object pose diffusion model is trained to generate plausible object pose during interactions given a human pose, our smoothness guidance sampling allows to generate plausible object motion for input sequential human poses.
In many cases, the assumption that the object trajectory should be smooth while HOI is valid, but in situations where the object vibrates within a small range (e.g., when drilling a hole with an electric drill) collding with other object, the assumption can be problematic.
To naturally model motions involving such collisions, physics information is required, and understanding such physics during HOI can be considered as potential future work.

\subsection{Modeling Dexterous Hand-Object Interaction}
\label{subsec:hand}
\begin{figure}[t]
\centering
\includegraphics[width=\columnwidth]{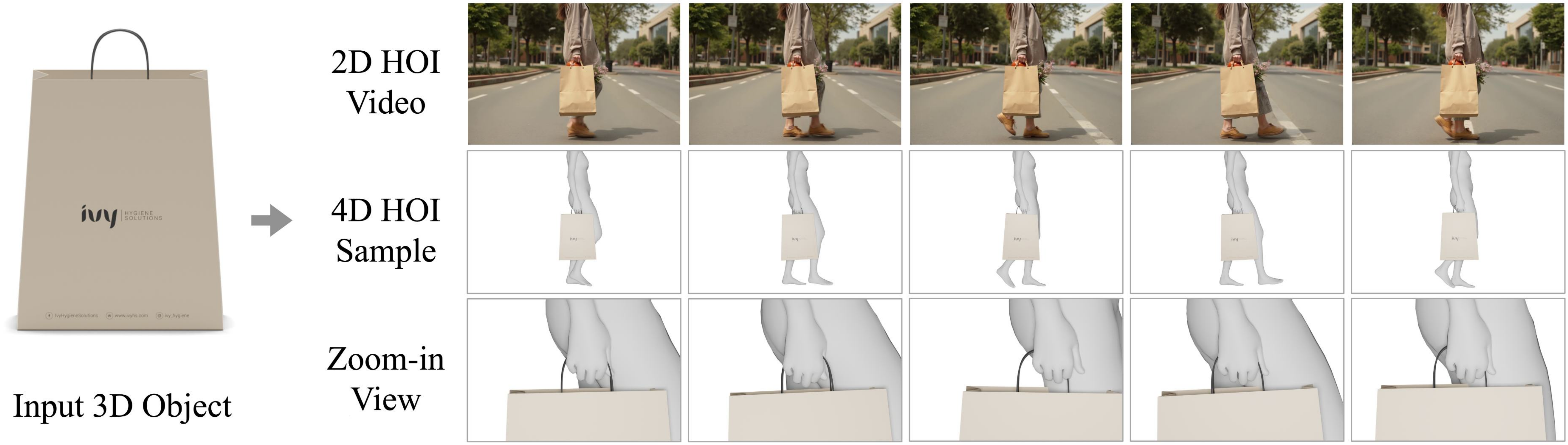}
\caption{\textbf{Hand Pose Extension.} DAVID uses SMPL-X as the human model, allowing hand pose extension in our 4D HOI samples.} 
\label{fig:hand_extension}
\end{figure}

Although we model the human with SMPL-X~\cite{SMPLX}, and recent 2D diffusion models demonstrate impressive quality in representing detailed hands, the pre-trained video diffusion model and 3D human estimator struggle to uplift the 2D hand from HOI images to high-quality 4D.
This hinders the modeling of dexterous hand-object interactions in both our 4D HOI samples and the learned Dynamic Affordance.
As a future direction, we can explore separately learning the hand patterns and merging them with the dynamic patterns we learned, with the expectation of improving the hand quality of the sampled HOI motion.
As shown in Fig.~\ref{fig:hand_extension}, we show that hand poses can be extended to our 4D HOI samples using a hand pose estimator~\cite{HaMeR} in a simple scenario.

\subsection{Concept Conflict}
\label{subsec:concept_conflict}
As we show that our LoRA~\cite{LoRA} has an advantage for modeling multiple concepts (\eg, combining existing knowledge of pre-trained model, and combining the knowledge of two individual LoRAs~\cite{LoRA}), the concept conflict may appear when combining two different concepts, similar to what occurs in image diffusion models.
When the two learned concepts show totally different human motion patterns (\eg, lifting a barbell, pushing a cart), we empirically observe that the result converges into two cases: (1) a motion is interpolated between two concepts, resulting implausible motion or (2) one motion is performed followed by the other.
Instead, when the two concepts are reasonably similar (\eg, holding an umbrella, riding a scooter), their motions can be combined to generate multi-object interactions.
However, we find that the relatively less coherent patterns (\eg, the position of the hand while riding a scooter) are removed while combining the concepts, which is the limitation of our application.

\end{document}